\documentclass[lettersize,journal]{IEEEtran}
\usepackage{amsmath,amsfonts}
\usepackage{algorithmic}
\usepackage{algorithm}
\usepackage{array}
\usepackage[caption=false,font=normalsize,labelfont=sf,textfont=sf]{subfig}
\usepackage{textcomp}
\usepackage{stfloats}
\usepackage{url}
\usepackage{verbatim}
\usepackage{graphicx}
\usepackage{cite}
\usepackage{booktabs}
\usepackage{makecell}
\usepackage{multirow}
\usepackage{multicol}
\usepackage{algorithm}
\usepackage{algorithmic}
\usepackage{setspace}
\usepackage{balance}
\usepackage[colorlinks,linkcolor=blue,anchorcolor=blue,citecolor=blue]{hyperref}

\hypersetup{
	colorlinks=true,
	linkcolor=red,
}

\begin{document}

\title{Confidence-Calibrated Face and Kinship Verification}

\author{Min Xu, Ximiao Zhang, and Xiuzhuang Zhou
        % <-this % stops a space
\thanks{This work was funded in part by the National Natural Science Foundation of China under Grants 62177034 and 61972046, and in part by the Beijing Natural Science Foundation under Grants 4202051. (Corresponding author: Xiuzhuang Zhou.)

Min Xu and Ximiao Zhang are with the College of Information and Engineering, Capital Normal University, Beijing 100048, China.
E-mail: xumin@cnu.edu.cn, 2211002048@cnu.edu.cn.

Xiuzhuang Zhou is with the School of Artificial Intelligence, Beijing University of Posts and Telecommunications, Beijing 100876, China. E-mail: xiuzhuang.zhou@bupt.edu.cn.}}

\markboth{Journal of \LaTeX\ Class Files,~Vol.~14, No.~8, August~2021}%
{Shell \MakeLowercase{\textit{et al.}}: A Sample Article Using IEEEtran.cls for IEEE Journals}

\maketitle

\begin{abstract}
    In this paper, we investigate the problem of prediction confidence in face and kinship verification. Most existing face and kinship verification methods focus on accuracy performance while ignoring confidence estimation for their prediction results. However, confidence estimation is essential for modeling reliability and trustworthiness in such high-risk tasks. To address this, we introduce an effective confidence measure that allows verification models to convert a similarity score into a confidence score for any given face pair. We further propose a confidence-calibrated approach, termed Angular Scaling Calibration (ASC). ASC is easy to implement and can be readily applied to existing verification models without model modifications, yielding accuracy-preserving and confidence-calibrated probabilistic verification models. In addition, we introduce the uncertainty in the calibrated confidence to boost the reliability and trustworthiness of the verification models in the presence of noisy data. To the best of our knowledge, our work presents the first comprehensive confidence-calibrated solution for modern face and kinship verification tasks. We conduct extensive experiments on four widely used face and kinship verification datasets, and the results demonstrate the effectiveness of our proposed approach. Code and models are available at \url{https://github.com/cnulab/ASC}.
\end{abstract}


\section{Introduction}
\label{Section1}

\IEEEPARstart{F}ace and kinship verification are two popular facial analysis tasks in computer vision. Face verification attempts to determine whether a given pair of face samples belongs to the same individual \cite{chopra2005learning}. Kinship verification aims to predict the existence of a kin relation between a given pair of face samples \cite{fang2010towards}. Over the past decade, face and kinship verification tasks have achieved remarkable advancements attributed to the application of deep learning technologies, particularly under controlled conditions. However, the two tasks in the wild are still challenging due to large intra-class variations, such as complex background, occlusion, and a variety of variations on illumination, pose and facial expression. Prior works on face and kinship verification are generally devoted to improve prediction accuracy, while paying little attention to confidence estimation. We contend that the reliability is also a key measure for evaluating the performance of these verification algorithms, and it becomes even more crucial for those verification systems deployed in high-risk scenarios. In this paper, we focus on modeling confidence estimation and calibration for face and kinship verification.

Accurate confidence estimation for face and kinship verification often presents a challenge. One reason for this can be the absence of labels that describe sample (pairs) uncertainty in most face or kinship verification datasets. To address this issue, existing methods employ quality networks \cite{liu2021eqface,xie2020inducing} or uncertainty estimation \cite{shi2019probabilistic,terhorst2020ser} to assist the decision-making. However, these approaches are more concerned with the quality of individual face images rather than the verification tasks. Additionally, modern neural networks tend to be over- or under-confident in their predictions. Hence, the estimated confidence of a face pair might not accurately reflect the verification accuracy. To overcome these challenges, we first propose a simple yet effective confidence measure for face and kinship verification. Our approach is inspired by the observation that if the pair similarity is close to the decision threshold, the model is less likely to make a correct prediction. Conversely, if the pair similarity is far from the threshold, the model's prediction is more likely to be correct. The proposed measure allows face and kinship verification models to convert the similarity score into a confidence score for a given face pair. We further develop a new algorithm to adjust the similarity of face pairs in angular space for confidence calibration, so that the calibrated confidence can well quantify the prediction confidence while maintaining the verification accuracy. Furthermore, we integrate the proposed confidence measure and confidence calibration methods with uncertainty estimation, which allows the confidence score to take into account both image quality and the verification process, providing a more reliable reference for decisions.

\begin{figure}[!t]
\centering
\includegraphics[width=\linewidth]{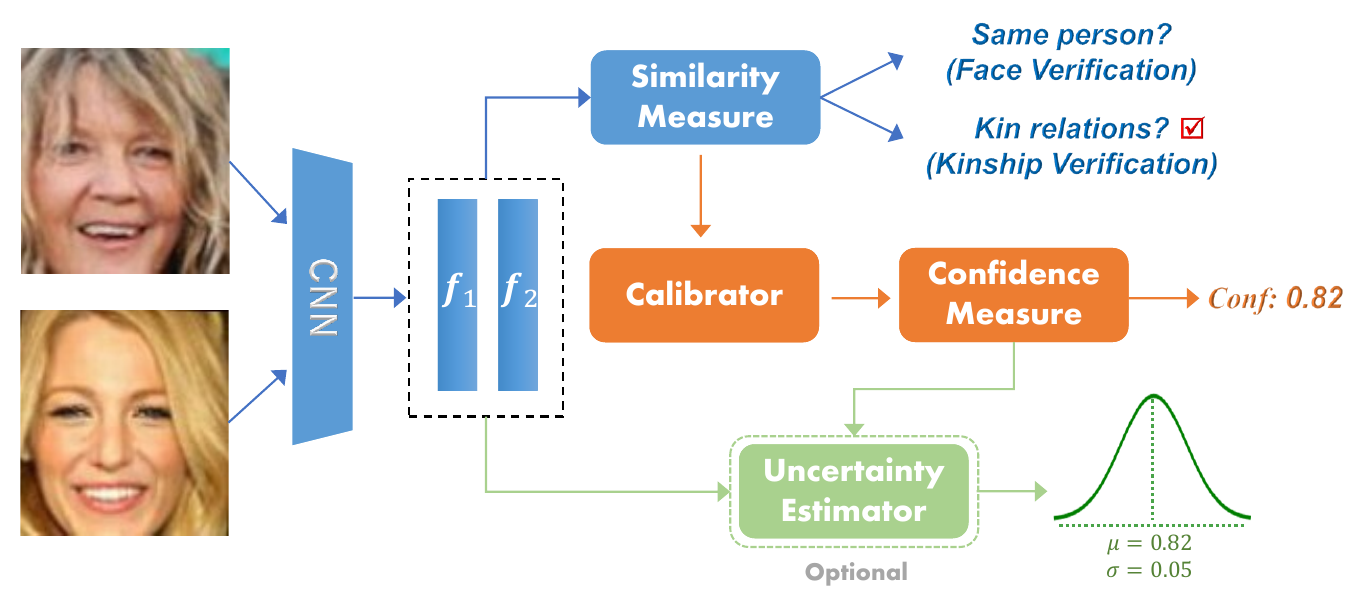}
\centering
\caption{The pipeline of our proposed confidence-calibrated face and kinship verification. Different from previous methods, our method gives well-calibrated confidence estimation on the prediction, by introducing a simple confidence measure and a flexible calibrator that can be directly applied to most existing face and kinship verification models without modifications. In addition, an uncertainty estimator can be optionally introduced to obtain a confidence score with uncertainty estimation, allowing for the assessment of the verification system's reliability and trustworthiness in the presence of data noise.}
\label{Fig1}
\end{figure}

The pipeline of our confidence-calibrated face and kinship verification is illustrated in Fig. \ref{Fig1}. The main contributions of this work can be outlined below:
\begin{itemize}
\item We introduce a new confidence measure for probabilistic face and kinship verification, allowing any off-the-shelf verification models to efficiently estimate the prediction confidence.
\item We develop a confidence-calibrated algorithm via angle scaling, which can be directly applied to existing face and kinship verification models without model modifications. To the best of our knowledge, this is the first general confidence-calibrated solution to face and kinship verification in a modern context.
\item We propose a method for estimating uncertainty in calibrated confidence to boost the reliability and trustworthiness of verification systems in the presence of noisy data.
\item Extensive experiments are conducted on both face and kinship datasets, and the results demonstrate that our proposed methods achieve state-of-the-art calibration and decision-aid performance.
\end{itemize}

The rest of the paper is organized as follows: Section \ref{Section2} first reviews prior works on face and kinship verification, and then discusses related works on confidence estimation in face analysis. In section \ref{Section3}, we elaborate our proposed confidence measure for face and kinship verification, followed by our confidence calibration and uncertainty estimation approaches. Section \ref{Section4} presents experimental results and analysis. Finally, we conclude the paper in section \ref{Section5}.

\section{Related Works}
\label{Section2}
\subsection{Face and Kinship Verification}
Face verification is a well-studied research topic in computer vision. In early studies, the low-level handcrafted features were used for verification and identification. The advent of deep neural networks in the past decade greatly enhances the performance of face verification tasks. Pioneering works such as DeepFace \cite{taigman2014deepface} and DeepID \cite{sun2015deepid3} introduced deep CNNs into face recognition. These works used deep models to learn effective high-level features, which significantly boosted face recognition performance. FaceNet \cite{schroff2015facenet} proposed the use of the triplet loss to learn a direct mapping from face images to a compact Euclidean space, where distances correspond directly to a measure of face similarity. Wen et al. \cite{wen2016discriminative} proposed a center loss to enhance the compactness of intra-class samples by minimizing the Euclidean distance between deep features and their corresponding class centers, while combining a softmax loss to guarantee inter-class differences. The introduction of Large-Margin Softmax \cite{liu2016large} sparked new ideas for face verification, and led to several notable studies, such as SpereFace \cite{liu2017sphereface}, CosFace \cite{wang2018cosface}, and ArcFace \cite{deng2019arcface}. These methods penalized the angles between deep features and their corresponding weights in angular space, effectively improving the discriminative ability of feature embeddings. Note that the angular-based distance measure is gradually replacing the Euclidean distance measure as the most popular method for face verification tasks, since the cosine of the angle and the softmax loss are inherently consistent.

The facial kinship verification problem dates back to the pioneering work by Fang et al. \cite{fang2010towards}. Since then, a variety of approaches have been proposed \cite{robinson2021survey,wu2022facial}. These approaches can be roughly divided into two categories: traditional methods based on feature or metric learning, and deep learning-based approaches developed in recent years. The former category seeks to learn a feature encoder or distance metric from pairwise kinship samples. A representative work of this line was the neighborhood repulsed metric learning (NRML) \cite{lu2013neighborhood}, which aimed to learn a distance metric that maximized the distance between negative pairs while minimizing the distance between positive pairs. Following this, \cite{yan2014discriminative,zhou2016ensemble,hu2019multi,xu2016kinship,zhou2016kinship,wang2017kinship} expanded on this concept by developing the novel approaches combining multiple features \cite{yan2014discriminative,xu2016kinship}, multiple views \cite{hu2019multi,xu2016kinship,zhou2016kinship}, multiple similarity \cite{zhou2016ensemble,zhou2016kinship}, and denoising \cite{wang2017kinship}. More recently, the traditional methods \cite{lu2017discriminative,qin2019social,duan2017face,wang2018cross,kohli2018deep,zhang2015kinship,zhou2019learning,dong2021kinship,hu2017local,wei2019adversarial,zhang2019deep,goyal2021robust} have been innovatively enhanced by incorporating deep learning techniques, such as deep metric learning, deep feature representation, or their combination to tackle the kinship recognition problem \cite{lu2017discriminative,wang2018cross,zhou2019learning,dong2021kinship}. Moreover, advanced kinship recognition solutions have emerged that leverage generative modeling \cite{zhu2021adversarial,ozkan2018kinshipgan,gao2021dna} or graph representation \cite{li2020graph} to further bolster the robustness of kinship verification.

As discussed above, most existing face and kinship verification methods focus on accuracy performance, while overlooking the importance of confidence estimation in their predictions. Even though a few attempts have been made recently to estimate the prediction confidence \cite{huber2022stating}, the estimated confidence can often be inaccurate due to the inadequate calibration of modern DNNs \cite{guo2017calibration}. Differently, our method provides well-calibrated prediction confidence while maintaining the verification accuracy.

\subsection{Uncertainty and Confidence Estimation in Face Analysis}

Uncertainty estimation has been integrated into numerous computer vision tasks in recent years to enhance system robustness and interpretability. These tasks range from object detection \cite{choi2019gaussian}, semantic segmentation \cite{kendall2015bayesian}, to image retrieval \cite{warburg2021bayesian,taha2019unsupervised}. Uncertainty is commonly classified into two types: aleatoric uncertainty and epistemic uncertainty \cite{kendall2017uncertainties}. Aleatoric uncertainty relates to the noise-related uncertainty in the data, whereas epistemic uncertainty refers to parameter uncertainty in the prediction model. Gal and Ghahramani \cite{gal2016dropout} suggested modeling predictive uncertainty by dropout training in deep neural networks as approximate Bayesian inference. Kendall and Gal \cite{kendall2017uncertainties} further developed a Bayesian deep learning framework that combines input-dependent aleatoric uncertainty with epistemic uncertainty. Additionally, Guo et al. \cite{guo2017calibration} conducted an investigation into various factors that influence the uncertainty of deep models and assessed the efficacy of several calibration methods for classification tasks.

PFE \cite{shi2019probabilistic} marked the initiation of integrating data uncertainty estimation into face recognition, achieving this by treating face embedding as a probabilistic embedding and determining the likelihood of a face pair belonging to the same identity. SER-FIQ \cite{terhorst2020ser} capitalized on the Bayesian approximation of model uncertainty using dropout as a mechanism to gauge image quality. Building on this, Xie et al. \cite{xie2020inducing} and Liu et al. \cite{liu2021eqface} put forth a method that involved training either an independent network or an auxiliary network branch to derive the quality scores of the images. The study represented in \cite{zhang2021relative} introduced a learning scheme for the relative uncertainty of a face pair via an extra network branch, with the obtained uncertainty acting as a weight to combine features from two distinct labels. Moreover, there have been concerted efforts to embed uncertainty analysis into face representation learning \cite{chang2020data,khan2019striking}.

Existing methods for modeling uncertainty in facial feature learning only provide an uncalibrated uncertainty estimation during the prediction stage, failing to accurately represent the confidence of the prediction. For verification tasks, our proposed confidence measure and calibration approach can provide a well-calibrated prediction confidence in the decision-making process, providing a direct representation of the prediction's probability of accuracy, thus aiding in risk assessment. The most closely related work to ours is the approach proposed in \cite{huber2022stating}, where uncertainty of the similarity score is estimated and propagated to the prediction confidence in face verification. However, this approach differs from our approach in the definition of the confidence measure, and it doesn't take into account the confidence calibration, leading to inaccurate confidence estimation. Experiment results on four widely-used datasets demonstrate the effectiveness of our proposed post-calibration method in face and kinship verification, particularly with respect to the uncertainty metric.

\section{Our Approach}
\label{Section3}
\subsection{Preliminaries: Face and Kinship Verification}

Given a face pair $(X,Y)$, we define the face embeddings as $x=f(X), y=f(Y)$, where $f(\cdot)$ is the feature embedding function often represented by modern DNNs. Existing face and kinship verification methods typically compute a cosine similarity, $s(x,y)$, as a similarity metric:
\begin{equation}
\label{equation 1}
s(x,y)=\frac{x \cdot y}{\vert\vert x\vert\vert \thinspace \vert\vert y \vert\vert}
\end{equation}

The decision function $g(s,\tau)$ for face and kinship verification can be defined by:
\begin{equation}
\label{equation 2}
g(s,\tau)=\left\{
\begin{aligned}
    1, & \qquad s \geq \tau \\
    -1, & \qquad s  <  \tau \\
\end{aligned}
\right.
\end{equation}
where $\tau$ is a predefined threshold that is often empirically set based on the Receiver Operating Characteristic (ROC) curve on a held-out validation set.

Recent works on face or kinship verification show that verification systems built even on popular DNNs may not work reliably, especially when the face images are partially occluded or of low resolution. Therefore, confidence estimation for face and kinship verification plays a key role in such a safety-critical task. However, most existing verification methods based on similarity measure fail to quantify the prediction confidence of face pairs, as the similarity score itself does not exactly reflect the prediction confidence. To address this issue, we propose a simple and flexible confidence measure to quantify the prediction confidence for any face and kinship verification models. Specifically, we estimate the prediction confidence based on the similarity $s$ and threshold $\tau$, and then calibrate the confidence in angular space so that the calibrated confidence is directly related to the probability of the prediction being correct.

\subsection{Confidence Measure}

Intuitively, if the similarity score of a face pair is equal to the left or right boundary values (-1 or 1), the prediction confidence will reach its maximum value of 1. Obviously, the prediction confidence relates not only to the similarity score but also to the decision threshold.

\begin{figure}[!t]
\centering
\includegraphics[width=\linewidth]{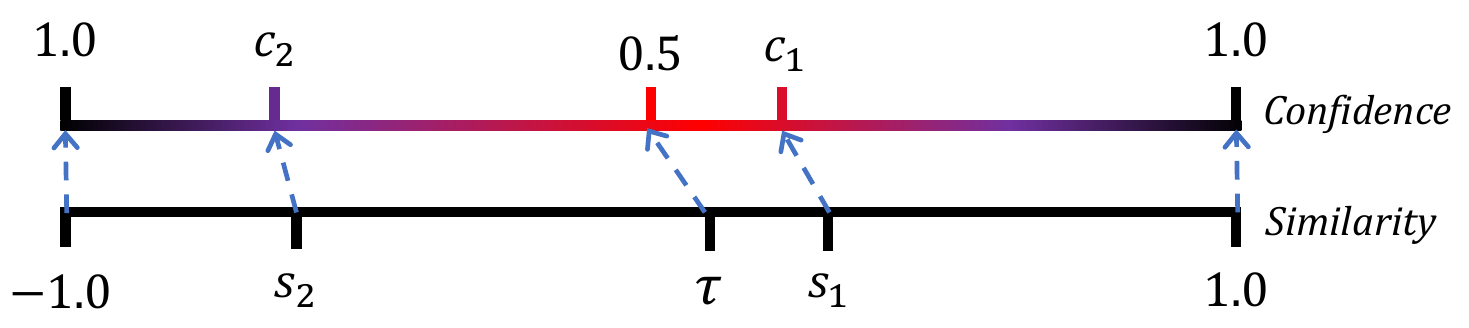}
\centering
\caption{An illustration for the relation between similarity score and prediction confidence. The closer to the decision threshold $\tau$ the similarity score $s$ is, the lower the confidence $c$ becomes.}
\label{Fig2}
\end{figure}

To model the relation between the similarity score and the prediction confidence for verification problems, we first define a confidence function $\varphi(s,\tau)$ based on the similarity $s$ and threshold $\tau$:
\begin{equation}
\label{equation 3}
\varphi(s,\tau)=\left\{
\begin{aligned}
    \frac{s-\tau}{1-\tau}, & \qquad g(s,\tau)=1 \\
    \frac{\tau-s}{1+\tau}, & \qquad g(s,\tau)=-1 \\
\end{aligned}
\right.
\end{equation}
where $s \in [-1,1]$, and $\tau \in (-1,1)$. The similarity threshold $\tau$ divides the cosine similarity interval $[-1,1]$ into positive and negative parts, with size of $1-\tau$ and $1+\tau$, respectively, as shown in Fig. \ref{Fig2}.

Note that face or kinship verification task is a binary decision problem. Hence, the probabilistic confidence $c$ for a prediction $g(s,\tau)$ based on Eq. \eqref{equation 2} should be greater than or equal to 0.5. Then we apply an affine transformation to $\varphi(s,\tau)$ to obtain $c(s,\tau)$:
\begin{equation}
\label{equation_4}
c(s,\tau)=\frac{1}{2}\varphi(s,\tau)+\frac{1}{2}
\end{equation}

As shown in Fig. \ref{Fig2}, if $s$ takes a value further away from $\tau$ (e.g., $s=s_2$), the prediction confidence $c$ becomes much higher (e.g., $c=c_2$). On the contrary, if the value of $s$ is closer to $\tau$ (e.g., $s=s_1$), the confidence $c$ decreases (e.g., $c=c_1$), indicating the more likely the model makes incorrect predictions. So far, we have proposed a flexible confidence measure that can be directly applied to any off-the-shelf verification models to yield a probabilistic estimation of prediction confidence.

\subsection{Confidence Calibration via Angular Scaling}

In reality, a bias exists between the verification accuracy and prediction confidence in face and kinship verification problems. This can be attributed to the fact that modern DNNs are often miscalibrated\cite{guo2017calibration}. As such, the proposed confidence measure may not produce a well-matched confidence to the expected accuracy of the model. For classification tasks, it is typical to calibrate the model by scaling the $logit$\cite{guo2017calibration}. However, if a similar process is applied to directly scale the similarity $s$ in the verification task, the similarity $s$, the threshold $\tau$, and the similarity interval will be equally scaled. This results in no change in the prediction confidence, according to Eq. \eqref{equation 3}. Inspired by the work of \cite{deng2019arcface,liu2017sphereface,wang2018cosface}, we propose a post-calibration method via angular scaling, which effectively calibrates the prediction confidence without the need for retraining the model.

We propose to calibrate the prediction confidence by calibrating the angle $\theta$ between pairs of faces on a recalibration dataset $D_c=\{(X_i,Y_i),Z_i\}_{i=1}^N$, as well as a feature encoder denoted as $f(\cdot)$. For a sample pair $(X_i,Y_i)$ with label $Z_i\in \{-1,1\}$ in $D_c$, its feature representation and cosine similarity are denoted by $x_i=f(X_i), y_i=f(Y_i)$, and $s_i=s(x_i,y_i)$, respectively. Our Angular Scaling Calibration (ASC) learns to adjust the angle between the face pairs based on the similarity set $S_c=\{s_i\}_{i=1}^N$ and its labels $Z_c=\{Z_i\}_{i=1}^N$, derived from the recalibration set $D_c$. We define our calibration function as follows:
\begin{equation}
\label{equation_5}
\psi(s)=cos(arccos(s)*w+b)
\end{equation}
where $w$ $(w>0)$ and $b$ are learnable scalar parameters. The calibrated similarity and threshold are given by $s'=\psi(s)$ and $\tau'=\psi(\tau)$, respectively. The calibrated angle $\theta'$ satisfies $\theta' \in [0,\pi]$, where $\theta'=arccos(s)*w+b$.

To learn the calibration parameter $w$ and $b$, we minimize the following objective function:
\begin{equation}
\label{equation_6}
\mathcal{L}_{ASC}=\frac{1}{N}\sum_{i=1}^{N}(\psi(s_i)-Z_{i})^2
\end{equation}
where $Z_{i}$ equals 1 if the pair $(X_i,Y_i)$ is a positive pair, and -1 otherwise.

The detail of the calibration procedure via angular scaling is summarized in Algorithm \ref{alg:alg1}. In the prediction stage, we use the calibrated similarity $s'$ and the calibrated threshold $\tau'$ to compute calibrated prediction confidence, in accordance with Eq. \eqref{equation_4}.

\begin{algorithm}[b]
\caption{Angular Scaling Calibration (ASC)}\label{alg:alg1}
\begin{algorithmic}
\STATE
\STATE \textbf{Input: }\text{Recalibration set} $D_c=\{(X_i,Y_i),Z_i\}_{i=1}^N$.
\STATE \textbf{Output: }\text{Calibration parameters} $w,b$.
\STATE \textbf{Set} $w=1,b=0$;
\STATE \text{Compute the similarity set $S_c$ of feature pairs}
\STATE \text{\quad\quad\quad $S_c=\{s(x_i,y_i)\}_{i=1}^N$;}
\STATE \textbf{while}\text{ not converged}\textbf{ do}
\STATE \hspace{0.2cm}\text{1. Update the similarity set according to Eq. \eqref{equation_5};}
\STATE \hspace{0.2cm}\text{2. Update $w$,$b$ by optimizing the objective \eqref{equation_6};}
\STATE\textbf{end}
\STATE \textbf{return}  $w,b$
\end{algorithmic}
\end{algorithm}

Now, we briefly explain how our ASC ensures preservation of prediction accuracy. Firstly, let $\tau$ denote the uncalibrated threshold, and we are given the positive pair $(X_p,Y_p)$ and negative pair $(X_n,Y_n)$, and their features are $(x_p,y_p)$ and $(x_n,y_n)$, respectively. It is clearly that $s(x_n,y_n) < \tau \leq s(x_p,y_p) $, and since $arccos$ decreases monotonically in the interval $[-1,1]$, we ascertain that $\theta_n > \theta_{\tau} \geq \theta_p$. We then use $w,b$ to adjust the angle between the two features. Considering that $w>0$, if follows that $\theta_n'>\theta_{\tau}' \ge \theta_{p}'$, with $\theta_{n}',\theta_{\tau}',\theta_{p}'$ are falling within $[0,\pi]$. Likewise, we can also conclude $cos(\theta_{n}')<cos(\theta_{\tau}') \leq cos(\theta_{p}')$. Hence, it can be concluded that $s'(x_n,y_n)<\tau' \leq s'(x_p,y_p) $. This indicates that our proposed ASC learns to calibrate the prediction confidence while maintaining the verification accuracy.

To evaluate the calibration performance of our proposed method for face and kinship verification, we employ the ECE (Expected Calibration Error)\cite{naeini2015obtaining} as our evaluation metric. The ECE is computed as the weighted average difference across bins between the expected accuracy and the prediction confidence, as illustrated in Eq. \eqref{equation_7}:
\begin{equation}
\label{equation_7}
ECE=\sum_{m=1}^{M}\frac{\vert B_m \vert}{n}\vert acc(B_m)-conf(B_m) \vert
\end{equation}
where $B_m$ represents all predictions fall in the $m$th bin, $M$ denotes the number of bins for partition of the confidence interval $[0.5,1]$, $n$ is the total number of samples, $acc(B_m)$ and $conf(B_m)$ are the accuracy and confidence of $m$th bin, respectively, as calculated by Eq. \eqref{equation_8} and Eq. \eqref{equation_9}:
\begin{equation}
\label{equation_8}
acc(B_m)=\frac{1}{\vert B_m \vert}\sum_{i \in B_m}1(g(s_i,\tau)=Z_i)
\end{equation}

\begin{equation}
\label{equation_9}
conf(B_m)=\frac{1}{\vert B_m \vert}\sum_{i \in B_m}c(s_i,\tau)
\end{equation}

We use equal-size interval partition \cite{nixon2019measuring}, that is, the size of each bin is equal to $1/2M$. Note that a ideal calibration typically exhibits a minimal ECE. Specifically, when the accuracy of each bin is equal to its confidence, the ECE of the model will be zero, indicating that it is perfectly calibrated.

\subsection{Uncertainty Estimation for Confidence Calibration}

Due to our proposed confidence measure and confidence calibration methods operating at the similarity score level, they do not refer to information about image quality. For instance, in a scenario where two images are markedly blurry, thereby lacking sufficient facial information, the system may still yield high prediction confidence for these images due to elevated similarity scores. This occurrence may introduce potential risks to decision-making systems. To address this issue, we integrate uncertainty estimation into our confidence measure and confidence calibration methods. This enables the integration of image quality estimation into prediction confidence while simultaneously maintaining the calibration of confidence.

We follow the approach of \cite{shi2019probabilistic}, employing probabilistic face embedding instead of point embedding. For a given face image $X$, we have a probabilistic embedding $\mathcal N(\mu_x,\sigma_x^{2}\textbf{I})$, where $\mu_x, \sigma_x^2 \in R^L$ correspondingly signify the mean and variance of various dimensions in the feature, and $L$ denotes the number of feature dimension. The center $\mu_x$ signifies the original image embedding and $\sigma_x^2$ is predicted via an additional network branch. To simplify computation, we assume that the original feature embedding is \texttt{L2} normalized and all dimensions are independent and identically distributed. The log likelihood of a face image pair $(X,Y)$ being attributed to the same individual (or having a kinship) is denoted as:
\begin{gather}
\label{equation_10}
\begin{align}
\log p(X=Y)=&-\frac{1}{2}\sum_{l=1}^{L}(\frac{(\mu_{x_l}-\mu_{y_l})^2}{\sigma_{x_l}^2-\sigma_{y_l}^2}+\log(\sigma_{x_l}^2-\sigma_{y_l}^2)) \notag\\
&-const
\end{align}
\end{gather}
where $const=\tfrac{L}{2}\log2\pi$, and $\mu_{x_l}, \sigma_{x_l}$ refer to the $l$th dimension of $\mu_{x}$ and $\sigma_{x}$.

We maximize the log likelihood to optimize the additional network branch for accurate uncertainty estimation. The loss function is:
\begin{equation}
\label{equation_11}
\mathcal{L}_{PFE}=\frac{1}{\vert \mathcal{P} \vert}\sum_{\scriptsize (X_i,Y_i) \in \mathcal{P}}-\log p(X_i=Y_i)
\end{equation}
where $\mathcal{P}$ is the set of all positive pairs in the training dataset. For a given pair of face images $(X, Y)$, with corresponding face embeddings $(x, y)$, and the learned data uncertainty $(\sigma_x^2, \sigma_y^2)$, the uncertainty of calibrated confidence can be derived by utilizing the uncertainty propagation \cite{ku1966notes}. This formulation is given as:
\begin{equation}
\label{equation_12}
\sigma_c^2(x, y, \sigma_x^2, \sigma_y^2)=\left\{
\begin{aligned}
    \frac{\sigma_s^2(x, y, \sigma_x^2, \sigma_y^2)}{4[1-\psi(\tau)]^2}, & \quad s(x,y) \geq \tau \\
    \frac{\sigma_s^2(x, y, \sigma_x^2, \sigma_y^2)}{4[1+\psi(\tau)]^2}, & \quad s(x,y) < \tau \\
\end{aligned}
\right.
\end{equation}
where
\begin{equation}
\label{equation_13}
\sigma_s^2(x,y,\sigma_x^2, \sigma_y^2)=\frac{w^2[1-\psi^2[s(x,y)]]}{1-s^2(x,y)}\sum_{l=1}^{L}(y_l^2\sigma_{x_l}^2+x_l^2\sigma_{y_l}^2)
\end{equation}
where $\psi(\cdot)$ denotes the calibration function defined in Eq. \eqref{equation_5}, $s(x,y)$ represents the cosine similarity function, which can be simplified as $s(x,y) = x \cdot y$ when $\vert\vert x \vert\vert,\vert\vert y \vert\vert=1$. Specifically, when the calibration parameters $w=1$ and $b=0$, the uncertainty of the uncalibrated confidence aligns with Eq. \eqref{equation_12}. Detailed derivations pertaining to Eq. \eqref{equation_12} and \eqref{equation_13} are available in the \nameref{appendix}.

By introducing uncertainty estimation, the prediction confidence, denoted as $c$, adheres to the normal distribution $\mathcal{N}(\mu_c,\sigma_c^{2})$. The center $\mu_c$ is obtained through the deterministic feature embedding following Eq. \eqref{equation_4}, which reflects the confidence of the verification process itself. And the variance $\sigma_c^2$ reflects the influence of the image quality on the prediction confidence. Image pairs of inferior quality exhibit a larger variance $\sigma_c^2$, while image pairs of superior quality demonstrate a smaller variance $\sigma_c^2$.

Intuitively, we can use $\tilde{\mathcal{C}}=\mu_c-\sigma_c$ as a substitute for $\mu_c$ to introduce uncertainty estimation in prediction confidence, which takes into account both image quality and the image similarity. However, $\tilde{\mathcal{C}}$ tends to lead to an under-confident prediction confidence, thereby resulting in miscalibration. Therefore, we propose a calibrated $\tilde{\mathcal{C}}$ as follows:
\begin{equation}
\label{equation_14}
\tilde{\mathcal{C}}=\mu_c-\alpha(\sigma_c-E(\sigma_c))
\end{equation}
where $E(\sigma_c)$ represents the mean of $\sigma_c$ for all pairs in the recalibration set, which reflects the average image quality. The hyperparameter $\alpha$ $(\alpha \geq 0)$ is the uncertainty weighting factor, and a larger $\alpha$ indicates that the prediction confidence places greater emphasis on the uncertainty estimation. Specifically, when $\alpha$ equals 0, the uncertainty estimation for the image is disregarded. Eq. \eqref{equation_14} ensures that the $\tilde{\mathcal{C}}$ fluctuates around $\mu_c$ and taking $E(\mu_c)$ as its mean value, thereby guaranteeing calibration performance while incorporating uncertainty estimation.

\section{Experiments}

\label{Section4}
To validate the efficacy of our proposed methods for face and kinship verification, we conduct extensive experiments on four face and kinship datasets, including FIW \cite{robinson2016families}, KinFaceW \cite{lu2013neighborhood}, LFW \cite{huang2008labeled}, and IJB-C \cite{maze2018iarpa}. In this section, we will report the accuracy, mean confidence, the ECE with $M$ equals 20 before and after calibration, as well as the performance of uncertainty estimation on these datasets. The experimental analysis will also be presented in detail.

\begin{table*}[tb]
\renewcommand\arraystretch{1.08}
\caption{ Accuracy (\%) and miscalibration (\%) performance of different models on FIW \cite{robinson2016families}. In each group, the best result is marked in red, and the next best result is marked in blue}
\label{tab:table1}
\centering
\resizebox{\linewidth}{!}{
\begin{tabular}{c|c|ccccccccccc|c|c}
\bottomrule
\multirow{2}{*}{Loss} & \multirow{2}{*}{Model} & \multicolumn {12}{c|}{Accuracy ($\uparrow$)} & \multirow{2}{*}{ECE ($\downarrow$)} \\
\cline{3-14}
&&SS & BB & \ SIBS & FD & MD & FS & MS & GFGD & GMGD & GFGS & GMGS & \textbf{AVG}  \\
\hline
\multirow{4}{*}{InfoNCE}&ResNet101&\textcolor{red}{82.84}&\textcolor{red}{82.06}&\textcolor{red}{80.32}&\textcolor{red}{76.69}&\textcolor{red}{80.59}&\textcolor{red}{82.81}&\textcolor{red}{76.47}
                                        &\textcolor{red}{78.10}&\textcolor{red}{71.38}&\textcolor{red}{71.02}&\textcolor{blue}{60.34}&\textcolor{red}{80.19}&22.85\\
                        &ResNet50 &\textcolor{blue}{79.00}&\textcolor{blue}{74.81}&\textcolor{blue}{75.92}&\textcolor{blue}{71.82}&\textcolor{blue}{74.15}&\textcolor{blue}{78.60}
                                            &\textcolor{blue}{72.10}&\textcolor{blue}{73.36}&\textcolor{blue}{70.26}&\textcolor{blue}{63.67}&\textcolor{red}{61.45}&\textcolor{blue}{75.01}&16.51\\
                        &ResNet34 &76.49&73.46&72.99&70.60&72.42&75.88&69.23&67.27&62.83&62.86&56.98&72.87&\textcolor{blue}{12.72}\\
                        &VGG16    &67.20&66.32&62.27&61.61&63.93&66.90&62.08&67.95&61.71&\textcolor{blue}{63.67}&56.98&64.68&\textcolor{red}{5.97}\\
\hline
\multirow{4}{*}{Triplet}&ResNet101&\textcolor{red}{80.49}&\textcolor{red}{79.50}&\textcolor{red}{78.79}&\textcolor{red}{73.34}&\textcolor{red}{79.04}&\textcolor{red}{80.53}
                                    &\textcolor{red}{74.34}&\textcolor{red}{77.88}&\textcolor{red}{77.32}&\textcolor{red}{68.98}&\textcolor{blue}{59.22}&\textcolor{red}{77.94}&20.93\\
                        &ResNet50 &\textcolor{blue}{78.63}&\textcolor{blue}{75.80}&\textcolor{blue}{75.92}&\textcolor{blue}{72.51}&\textcolor{blue}{74.05}&\textcolor{blue}{76.49}
                        &67.17&68.40&66.17&\textcolor{blue}{66.94}&55.31&\textcolor{blue}{74.18}&16.97\\
                        &ResNet34 &75.17&73.95&71.29&72.26&71.03&76.26
                        &\textcolor{blue}{69.39}&\textcolor{blue}{69.75}&64.68&62.04&\textcolor{red}{59.78}&72.86&\textcolor{blue}{10.46}\\
                        &VGG16    &66.14&64.80&64.32&63.76&62.13&67.01
                        &64.02&65.69&\textcolor{blue}{68.03}&66.53&50.84&64.65&\textcolor{red}{7.87}\\
\hline
\multirow{4}{*}{ArcFace}&ResNet101&\textcolor{red}{80.23}&\textcolor{red}{79.36}&\textcolor{red}{79.38}&\textcolor{red}{75.62}&\textcolor{red}{77.52}&\textcolor{red}{81.19}
                                    &\textcolor{red}{73.54}&\textcolor{red}{76.75}&\textcolor{red}{75.84}&\textcolor{blue}{71.02}&\textcolor{blue}{61.45}&\textcolor{red}{78.02}&19.39\\
                        &ResNet50 &\textcolor{blue}{76.53}&73.13&\textcolor{blue}{73.23}&\textcolor{blue}{72.79}&\textcolor{blue}{73.63}&\textcolor{blue}{78.20}
                                    &\textcolor{blue}{70.11}&\textcolor{blue}{65.91}&57.25&\textcolor{red}{71.43}&\textcolor{red}{63.13}&\textcolor{blue}{73.87}&16.00\\
                        &ResNet34 &74.67&\textcolor{blue}{74.06}&70.94&71.12&71.00&75.66
                                    &67.25&64.55&62.83&58.37&\textcolor{blue}{61.45}&72.16&\textcolor{blue}{10.04}\\
                        &VGG16    &67.98&67.11&61.86&60.86&63.93&68.00
                                    &61.60&61.63&\textcolor{blue}{65.43}&55.92&58.10&64.83&\textcolor{red}{7.69}\\
\hline
\multirow{4}{*}{Softmax}&ResNet101&\textcolor{red}{78.51}&\textcolor{red}{76.60}&\textcolor{red}{75.57}&\textcolor{red}{75.48}&\textcolor{red}{76.64}&\textcolor{red}{81.28}
                                        &\textcolor{red}{76.57}&\textcolor{red}{73.81}&\textcolor{blue}{67.66}&\textcolor{blue}{68.98}&\textcolor{red}{65.92}&\textcolor{red}{77.25}&16.24\\
                        &ResNet50 &75.90&\textcolor{blue}{75.21}&\textcolor{blue}{73.29}&69.57&\textcolor{blue}{73.49}&\textcolor{blue}{78.06}
                                        &\textcolor{blue}{70.48}&\textcolor{blue}{66.37}&60.97&\textcolor{red}{77.14}&\textcolor{blue}{64.25}&\textcolor{blue}{73.74}&15.77\\
                        &ResNet34 &\textcolor{blue}{76.34}&74.18&72.47&\textcolor{blue}{70.76}&72.76&76.75
                                        &68.43&65.46&64.31&60.00&59.22&73.06&\textcolor{blue}{8.90}\\
                        &VGG16    &67.76&67.80&61.45&61.75&64.35&66.32
                                        &60.59&60.05&\textcolor{red}{68.40}&56.73&56.42&64.71&\textcolor{red}{5.40}\\
\toprule
\end{tabular}
}
\end{table*}

\subsection{Datasets and Experimental Settings}
\textsl{1) FIW} \cite{robinson2016families}: FIW (Families In the Wild) is the largest visual kinship recognition dataset to date, consisting of over 13,000 facial images from 1000 families, with 11 pairwise kin relationships that can be divided into three sub-groups: siblings type (i.e., sister-sister, brother-brother, and sister-brother), parent-child type (i.e., father-son, father-daughter, mother-son, and mother-daughter), and grandparent-grandchild type (i.e., grandfather-grandson, grandfather-granddaughter, grandmother-grandson, and grandmother-granddaughter). The FIW is a challenging dataset because all the images are captured in the wild with partial occlusion and large variations in background, pose, expression, illumination, and so on.

InfoNCE \cite{chen2020simple}, Triplet \cite{schroff2015facenet}, Softmax \cite{taigman2014deepface}, and ArcFace \cite{deng2019arcface} are four widely-used loss functions in face and kinship verification tasks. We first use the four loss functions to train models in FIW \cite{robinson2016families} with different backbones pre-trained on the MS1MV2 \cite{deng2019arcface}. We follow the data partitioning and evaluation protocol of RFIW 2021 \cite{robinson20215th}. Moreover, we employ RetinaFace \cite{deng2020retinaface} for face alignment and the images are cropped into $112\times112$ pixels.

\textsl{2) KinFaceW} \cite{lu2013neighborhood}: The KinFaceW-I and KinFaceW-II datasets are widely used kinship datasets comprised of Internet-collected images of public figures and celebrities. The two datasets contain four kin relations: Father-Son, Father-Daughter, Mother-Son, and Mother-Daughter. For these four kin relations, KinFaceW-I has 156, 134, 116, and 127 pairs of images with kin relationship. In KinFaceW-II, there are 250 pairs of images for each relationship. KinFaceW-I differs from KinFaceW-II in that, in most cases, the face pairs in the former dataset are from distinct photos while those in the latter one are from the same photos. Each image in the two datasets is aligned and resized into $64\times64$ for feature extraction.

We evaluate two traditional and one deep-learning kinship verification methods on the KinFaceW datasets, including: 1) NRML (LBP) \cite{lu2013neighborhood}: for each face image, we extract 3776-dimensional uniform pattern LBP features to kinship verification. 2) NRML (HOG) \cite{lu2013neighborhood}: first, we partition each image into non-overlapping blocks. Then, we extract a 9-dimensional HOG feature for each block and concatenate them to generate a 2880-dimensional feature. 3) InfoNCE (ResNet18) \cite{chen2020simple}: we use InfoNCE loss \cite{chen2020simple} to fine-tune a ResNet18 model pre-trained on the MS1MV2 \cite{deng2019arcface}. In this experiment, we use a five-fold cross-validation strategy in an image-unrestricted setting.

\textsl{3) LFW} \cite{huang2008labeled}: LFW contains 13,233 face images collected from 5,749 individuals, of which 1,680 have 2 or more face images.

We use CASIA-WebFace \cite{yi2014learning} as the training set to train different backbones with the ArcFace loss \cite{deng2019arcface}, and evaluate the verification performance of these models on the LFW dataset. We calibrate the confidence of different backbones using ASC in the 10-fold cross-validation scheme.

\textsl{4) IJB-C} \cite{maze2018iarpa}: IJB-C (IARPA Janus Benchmark-C) dataset contains 31,334 images and 11,779 videos captured under unconstrained condition from 3,531 subjects. There are a total of 23,124 templates, with 19,557 positive matches and 15,638,932 negative matches to enable performance evaluations at low FAR.

In this experiment, we adopt the 1:1 Mixed Verification protocol \cite{maze2018iarpa}, that is, a single feature vector is created by taking a weighted average of the frames in each template, so that all frames in a video belonging to the same subject have the same cumulative weight as a single still image. We perform face verification on the IJB-C dataset using different backbones trained on the MS1MV2 \cite{deng2019arcface} with the ArcFace loss \cite{deng2019arcface} and the Triplet loss \cite{schroff2015facenet}. To perform confidence estimation and calibration, we use five-fold cross-validation with stratified sampling of positive and negative samples, with four of the folds being the recalibration set and the remaining one being the test set.

\begin{figure*}[t]
\begin{minipage}[t]{0.47\textwidth}
        \vspace{0pt}
        \centering
        \makeatletter\def\@captype{figure}\makeatother
        \includegraphics[width=\linewidth]{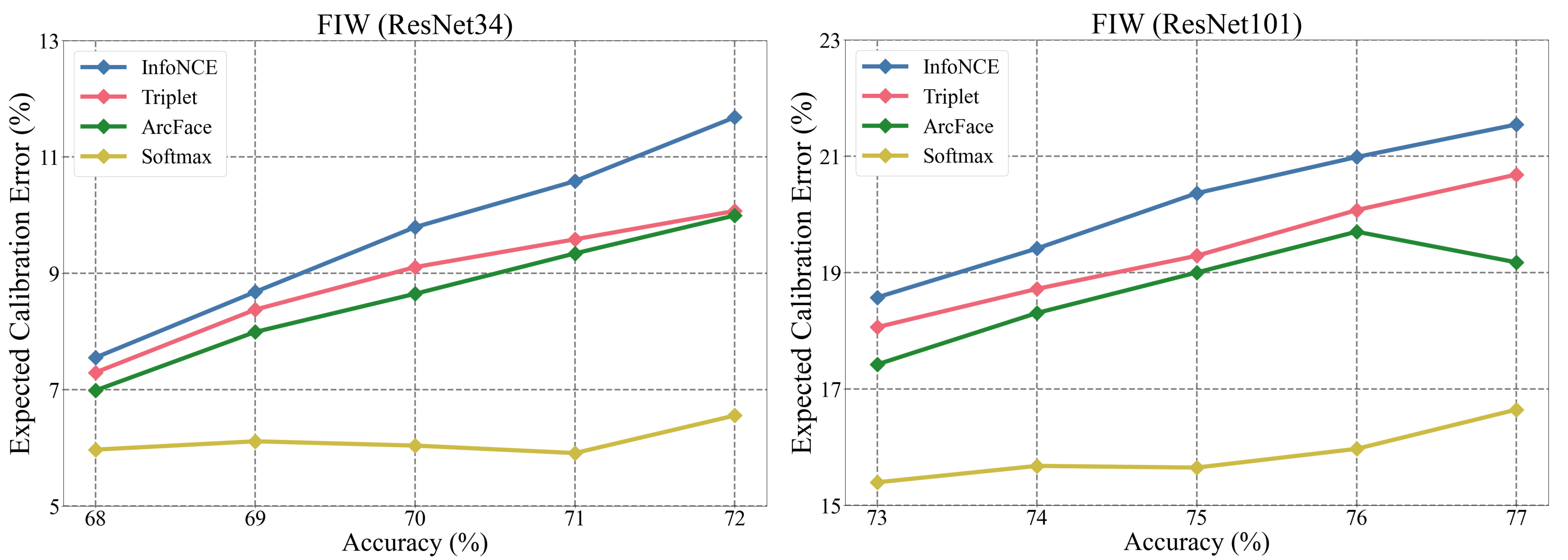}
        \caption{The trend of ECE (\%) vs. accuracy (\%) of different losses during training on FIW \cite{robinson2016families}.}
        \label{Fig3}
        \vspace{8pt}
        \hspace*{\fill}
        \includegraphics[width=\linewidth]{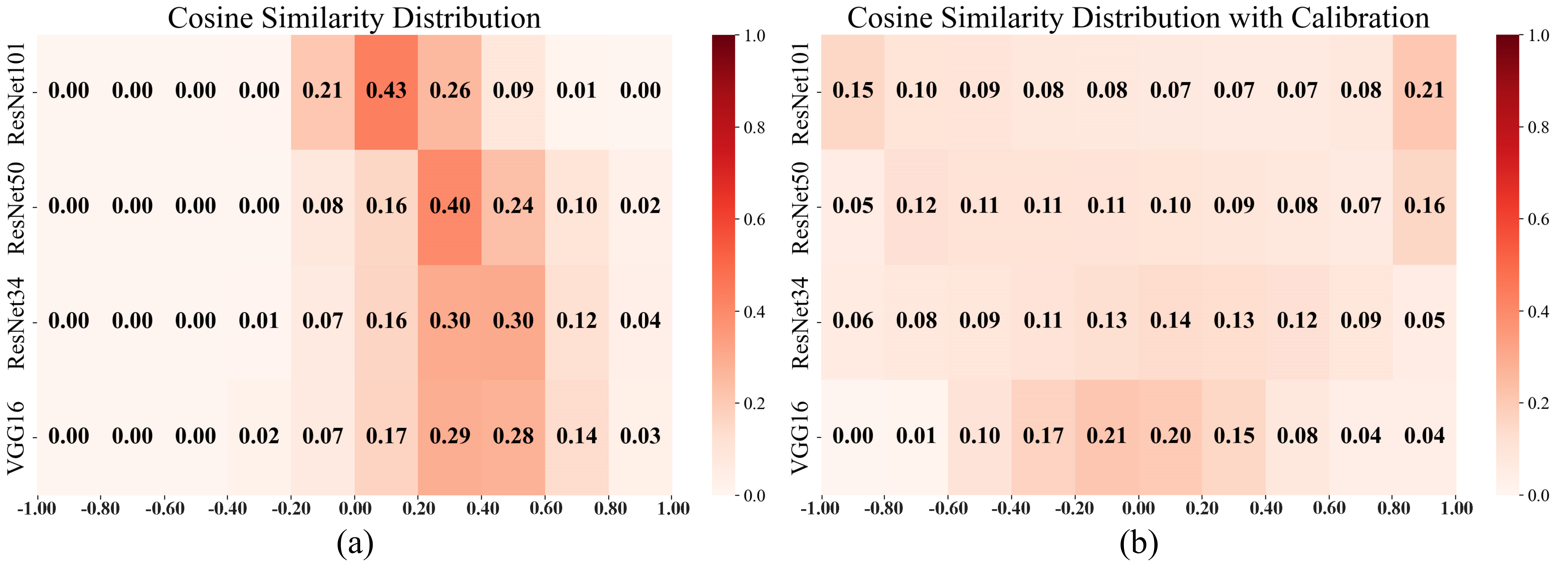}
       \caption{Similarity distributions before calibration (a) and after calibration with ASC (b). From top to bottom are results of ResNet101, ResNet50, ResNet34, and VGG16, respectively.}
        \label{Fig4}

\end{minipage}\quad
\begin{minipage}[t]{0.50\textwidth}
        \vspace{0pt}
		\centering
        \includegraphics[width=\textwidth]{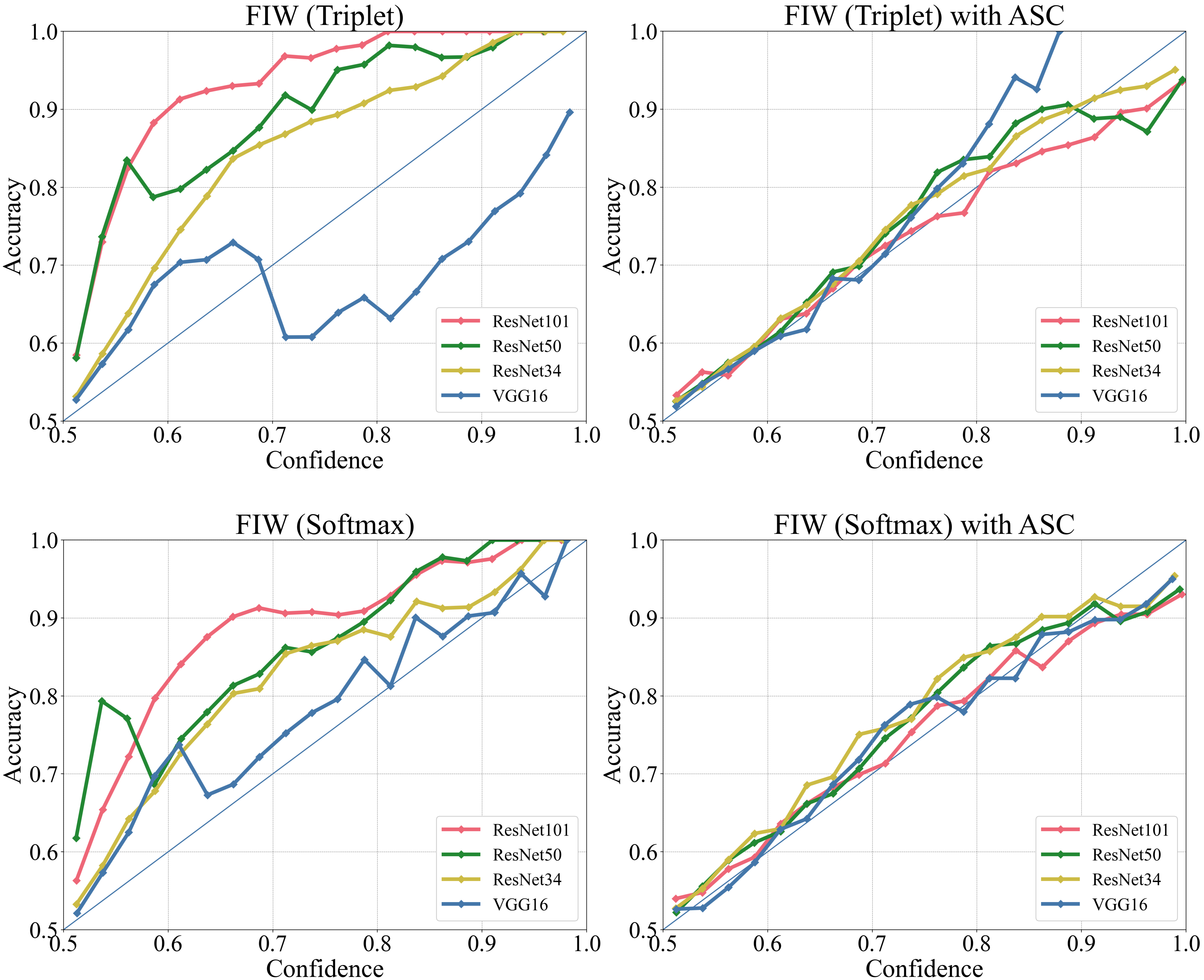}
        \centering
        \caption{Calibration curves of different models before and after calibration (with ASC) on FIW \cite{robinson2016families}.}
        \label{Fig5}
\end{minipage}
\end{figure*}

\subsection{Experimental Results on FIW}
In Table \ref{tab:table1}, we show the verification accuracy, the weighted average accuracy, and the ECE on FIW dataset \cite{robinson2016families} for different models and loss functions. From these results, we make the observations as below:

\begin{enumerate}
\item For all four loss functions, models with deeper architecture exhibit superior feature representation abilities. However, we also observe that the increase in depth leads to a negative influence on model calibration.
\item Compared with Softmax \cite{taigman2014deepface} and ArcFace \cite{deng2019arcface} losses, InfoNCE \cite{chen2020simple} and Triplet \cite{schroff2015facenet} losses achieve certain performance enhancements, however, it also comes with an increase in calibration error.
\item ArcFace loss \cite{deng2019arcface} leverages the angular margin penalty to enforce extra intra-class compactness and inter-class discrepancy, achieving superior verification performance with larger calibration error than Softmax loss \cite{taigman2014deepface}.
\end{enumerate}

In Fig. \ref{Fig3}, we plot the calibration-accuracy during the training process of ResNet34 and ResNet101, intuitively showing the trend of the model's ECE and verification accuracy with four different loss functions. It can be seen that the ECE of both models (ResNet34 and ResNet101) generally increase with the verification accuracy of the models. Also, if the model employs a loss directly optimizing the feature representation, such as InfoNCE \cite{chen2020simple} or Triplet \cite{schroff2015facenet}, the model exhibits a larger ECE with the same accuracy.

In addition, we conduct experiments on the FIW dataset \cite{robinson2016families} to evaluate the effects of the ASC. Table \ref{tab:table2} shows the ECE and the threshold values before and after calibration. Specifically, we adopt LBFGS\cite{nocedal1980updating} as our calibration optimizer with a learning rate of 0.01 and setting an upper limit of 1000 iterations. The calibrated threshold and calibration parameters ($w$ and $b$) are used to evaluate the ECE on the test data.

\begin{table}[!tbp]
\renewcommand\arraystretch{1.18}
\caption{ ECE (\%) and threshold of different models before and after calibration (with ASC) on FIW \cite{robinson2016families}}
\label{tab:table2}
\centering
\resizebox{0.9\linewidth}{!}{
\begin{tabular}{c|c|c|c|c|c}
\bottomrule
    Loss & Model& ECE & ECE w/ ASC & {\Large $\tau$} & {\Large $\tau$} w/ ASC \\
    \hline
    \multirow{4}{*}{InfoNCE}&ResNet101&22.85&2.08&0.13&-0.01\\
                            &ResNet50&16.51&1.81&0.32&-0.03\\
                            &ResNet34&12.72&2.11&0.37&0.02\\
                            &VGG16   &\textbf{5.97}&\textbf{1.39}&0.39&0.01\\
    \hline
    \multirow{4}{*}{Triplet}&ResNet101&20.93&2.83&0.11&0.02\\
                            &ResNet50&16.97&2.83&0.32&-0.07\\
                            &ResNet34&10.46&1.87&0.26&0.06\\
                            &VGG16&\textbf{7.87}&\textbf{1.54}&0.71&0.08\\
    \hline
    \multirow{4}{*}{ArcFace}&ResNet101&19.39&2.62&0.13&0.07\\
                            &ResNet50&16.00&\textbf{2.45}&0.39&-0.09\\
                            &ResNet34&10.04&3.40&0.14&-0.02\\
                            &VGG16&\textbf{7.69}&2.90&0.20&-0.07\\
    \hline
    \multirow{4}{*}{Softmax}&ResNet101&16.24&2.79&0.13&-0.04\\
                            &ResNet50&15.77&2.80&0.64&-0.05\\
                            &ResNet34&8.90&3.56&0.17&-0.01\\
                            &VGG16&\textbf{5.40}&\textbf{1.81}&0.37&0.01\\
\toprule
\end{tabular}
}
\end{table}

From Table \ref{tab:table2}, we observe that our ASC achieves good calibration on different models (loss functions). Additionally, it is noted that the calibrated threshold tends to converge towards 0, contributing to an equitable distribution of positive and negative pairings within the cosine similarity interval.

Fig. \ref{Fig4} plots the cosine similarity distribution for different models with InfoNCE loss \cite{chen2020simple} before and after calibration on the FIW dataset \cite{robinson2016families}. We observe that the similarity before calibration is condensed within a smaller interval, leading to suboptimal calibration, whereas the similarity after calibration is distributed uniformly across different similarity levels.

Fig. \ref{Fig5} presents accuracy-confidence plots corresponding to different models with Triplet loss \cite{schroff2015facenet} and Softmax loss \cite{taigman2014deepface} on the FIW dataset \cite{robinson2016families} before and after calibration. The plots underscore the proficiency of the prediction confidence calibration via ASC applied to all models, where ideal calibration aligns with the line $y=x$. Evidently, our ASC proves effective in achieving well-calibrated predictions, as the recalibrated confidence accurately mirrors the prediction accuracy.

Table \ref{tab:table3} summarizes the average verification accuracy, average confidence before and after calibration of different models on three sub-groups of FIW \cite{robinson2016families}. It can be seen that the confidence before calibration is significantly lower or higher than the actual verification accuracy, while the recalibrated confidence aligns more closely with the ground truth. Therefore, verification models calibrated by ASC can provide a reliable confidence estimation to support decision-making in face and kinship verification systems.

\subsection{Experimental Results on KinFaceW}

In Table \ref{tab:table4}, we report the accuracy, the mean confidence, and the ECE before and after calibration on KinFaceW-I and KinFaceW-II datasets \cite{lu2013neighborhood}. Our findings reveal that kinship verification methods concerning same-gender relationships (FS, MD) outperform those of different-gender relationships (FD, MS) in terms of verification accuracy and confidence. However, the increment in accuracy proves to be more pronounced. For NRML \cite{lu2013neighborhood}, the HOG exhibits superior verification accuracy and confidence relative to LBP, while it has a larger ECE. Compared to traditional verification methods, deep-learning methods demonstrate less confidence, with the lowest confidence and the highest ECE on most kin relationships. Besides, the accuracy is randomly distributed over the bins due to the small size of the KinFaceW dataset \cite{lu2013neighborhood} and limited number of data samples in each bin. However, the methods calibrated by our ASC consistently show improved calibration performance in terms of both ECE and mean ECE.

\begin{table*}[!htbp]
\renewcommand\arraystretch{1.08}
\caption{ Accuracy (\%) and confidence (\%) of different models before and after calibration (with ASC) on different age groups of FIW \cite{robinson2016families}}
\tiny
\label{tab:table3}
\centering
\resizebox{0.95\linewidth}{!}{
\begin{tabular}{c|c|c|c|c|c|c|c|c|c}
\hline
\multirow{2}{*}{Model} & \multicolumn {3}{c|}{BB, SS, SIBS}& \multicolumn {3}{c|}{FS, MS, FD, MD} & \multicolumn {3}{c}{GFGS, GMGS, GFGD, GMGD} \\
\cline{2-10}
&Conf.& Conf. w/ ASC & Acc. & Conf. & Conf. w/ ASC & Acc. & Conf. & Conf. w/ ASC & Acc. \\
\hline
ResNet101&58.20&81.98&\textbf{79.80}&56.71&78.50&\textbf{77.01}&55.78&75.25&\textbf{72.89}\\
ResNet50 &59.02&75.83&76.91&56.41&71.68&72.89&54.49&67.66&65.49\\
ResNet34 &63.10&72.88&74.12&62.02&71.09&72.44&60.31&66.52&65.32\\
VGG16    &70.34&64.11&65.26&69.20&63.63&64.29&67.86&63.22&64.08\\
\hline
\end{tabular}
}
\end{table*}

\begin{table*}[!htbp]
\renewcommand\arraystretch{1.1}
\caption{ Accuracy (\%), confidence (\%), and ECE (\%) of different methods before and after calibration (with ASC) on KinFaceW \cite{lu2013neighborhood}}
\label{tab:table4}
\centering
\resizebox{0.99\linewidth}{!}{
\begin{tabular}{c|c|c|c|c|c|c|c|c|c|c|c|c|c|c|c|c|c}
\bottomrule
\multirow{2}{*}{Dataset} & \multicolumn {2}{c|}{\multirow{2}{*}{Methods}} & \multicolumn {3}{c|}{FS} & \multicolumn {3}{c|}{FD} & \multicolumn {3}{c|}{MS} & \multicolumn {3}{c|}{MD} & \multicolumn {3}{c}{Mean} \\
\cline{4-18}
&\multicolumn {2}{c|}{}& Acc. ($\uparrow$) & Conf. & ECE ($\downarrow$)& Acc. ($\uparrow$) & Conf. & ECE ($\downarrow$)& Acc. ($\uparrow$)& Conf. & ECE ($\downarrow$)& Acc. ($\uparrow$)& Conf. & ECE ($\downarrow$)& Acc. ($\uparrow$)& Conf. & ECE ($\downarrow$)\\
\hline
\multirow{6}{*}{KFW-I} & \multirow{2}{*}{\makecell{NRML\\(LBP)}} & w/o ASC
&\multirow{2}{*}{81.43}&57.01&27.03&\multirow{2}{*}{69.42}&55.40&19.24&\multirow{2}{*}{67.23}&55.94&18.04&\multirow{2}{*}{72.87}&56.08&19.48&\multirow{2}{*}{72.74}&56.11&20.95 \\
\cline{3-3}\cline{5-6}\cline{8-9}\cline{11-12}\cline{14-15}\cline{17-18}
&& w/ ASC &\multirow{2}{*}{}&79.94&1.97&\multirow{2}{*}{}&68.43&\textbf{1.39}&\multirow{2}{*}{}&64.64&2.49&\multirow{2}{*}{}&71.05&2.70&\multirow{2}{*}{}&71.02&2.14 \\
\cline{2-18}
& \multirow{2}{*}{\makecell{NRML\\(HOG)}} & w/o ASC
&\multirow{2}{*}{\textbf{83.68}}&57.52&25.91&\multirow{2}{*}{74.64}&56.88&19.35&\multirow{2}{*}{71.56}&56.32&20.26&\multirow{2}{*}{79.96}&57.16&23.69&\multirow{2}{*}{77.46}&56.97&22.30 \\
\cline{3-3}\cline{5-6}\cline{8-9}\cline{11-12}\cline{14-15}\cline{17-18}
&& w/ ASC &\multirow{2}{*}{}&81.10&3.24&\multirow{2}{*}{}&72.20&4.19&\multirow{2}{*}{}&68.40&3.98&\multirow{2}{*}{}&76.02&6.95&\multirow{2}{*}{}&74.43&4.59 \\
\cline{2-18}
& \multirow{2}{*}{\makecell{InfoNCE\\(ResNet18)}} & w/o ASC
&\multirow{2}{*}{83.34}&55.24&27.29&\multirow{2}{*}{\textbf{82.88}}&54.99&27.75&\multirow{2}{*}{\textbf{81.01}}&54.86&25.17&\multirow{2}{*}{\textbf{85.04}}&55.93&28.31&\multirow{2}{*}{\textbf{83.07}}&55.26&27.13 \\
\cline{3-3}\cline{5-6}\cline{8-9}\cline{11-12}\cline{14-15}\cline{17-18}
&& w/ ASC &\multirow{2}{*}{}&82.03&\textbf{1.47}&\multirow{2}{*}{}&81.50&1.86&\multirow{2}{*}{}&79.13&\textbf{1.97}&\multirow{2}{*}{}&83.91&\textbf{1.25}&\multirow{2}{*}{}&81.64&\textbf{1.64} \\
\hline
\multirow{6}{*}{KFW-II} & \multirow{2}{*}{\makecell{NRML\\(LBP)}} & w/o ASC
&\multirow{2}{*}{79.20}&54.96&24.16&\multirow{2}{*}{71.60}&54.53&18.34&\multirow{2}{*}{72.20}&54.51&18.90&\multirow{2}{*}{68.40}&54.70&16.10&\multirow{2}{*}{72.85}&54.68&19.38 \\
\cline{3-3}\cline{5-6}\cline{8-9}\cline{11-12}\cline{14-15}\cline{17-18}
&& w/ ASC &\multirow{2}{*}{}&78.65&1.22&\multirow{2}{*}{}&70.88&1.57&\multirow{2}{*}{}&71.82&1.67&\multirow{2}{*}{}&67.25&1.67&\multirow{2}{*}{}&72.15&1.53 \\
\cline{2-18}
& \multirow{2}{*}{\makecell{NRML\\(HOG)}} & w/o ASC
&\multirow{2}{*}{80.80}&55.75&24.90&\multirow{2}{*}{72.80}&55.14&19.16&\multirow{2}{*}{74.80}&55.41&20.57&\multirow{2}{*}{70.40}&55.24&18.48&\multirow{2}{*}{74.70}&55.39&20.78 \\
\cline{3-3}\cline{5-6}\cline{8-9}\cline{11-12}\cline{14-15}\cline{17-18}
&& w/ ASC &\multirow{2}{*}{}&79.76&1.06&\multirow{2}{*}{}&72.24&1.86&\multirow{2}{*}{}&74.27&1.68&\multirow{2}{*}{}&68.82&1.89&\multirow{2}{*}{}&73.77&1.62 \\
\cline{2-18}
& \multirow{2}{*}{\makecell{InfoNCE\\(ResNet18)}} & w/o ASC
&\multirow{2}{*}{\textbf{84.00}}&55.37&28.13&\multirow{2}{*}{\textbf{78.20}}&54.50&23.20&\multirow{2}{*}{\textbf{82.20}}&55.15&26.55&\multirow{2}{*}{\textbf{84.00}}&55.16&28.34&\multirow{2}{*}{\textbf{82.10}}&55.05&26.56  \\
\cline{3-3}\cline{5-6}\cline{8-9}\cline{11-12}\cline{14-15}\cline{17-18}
&& w/ ASC &\multirow{2}{*}{}&83.48&\textbf{0.60}&\multirow{2}{*}{}&78.16&\textbf{0.81}&\multirow{2}{*}{}&80.55&\textbf{1.54}&\multirow{2}{*}{}&83.17&\textbf{1.06}&\multirow{2}{*}{}&81.34&\textbf{1.00}  \\
\toprule
\end{tabular}
}
\end{table*}

\begin{figure*}[!htbp]
        \centering
\begin{minipage}[t]{0.48\textwidth}
        \vspace{0pt}
        \centering
        \makeatletter\def\@captype{figure}\makeatother
        \includegraphics[width=\linewidth]{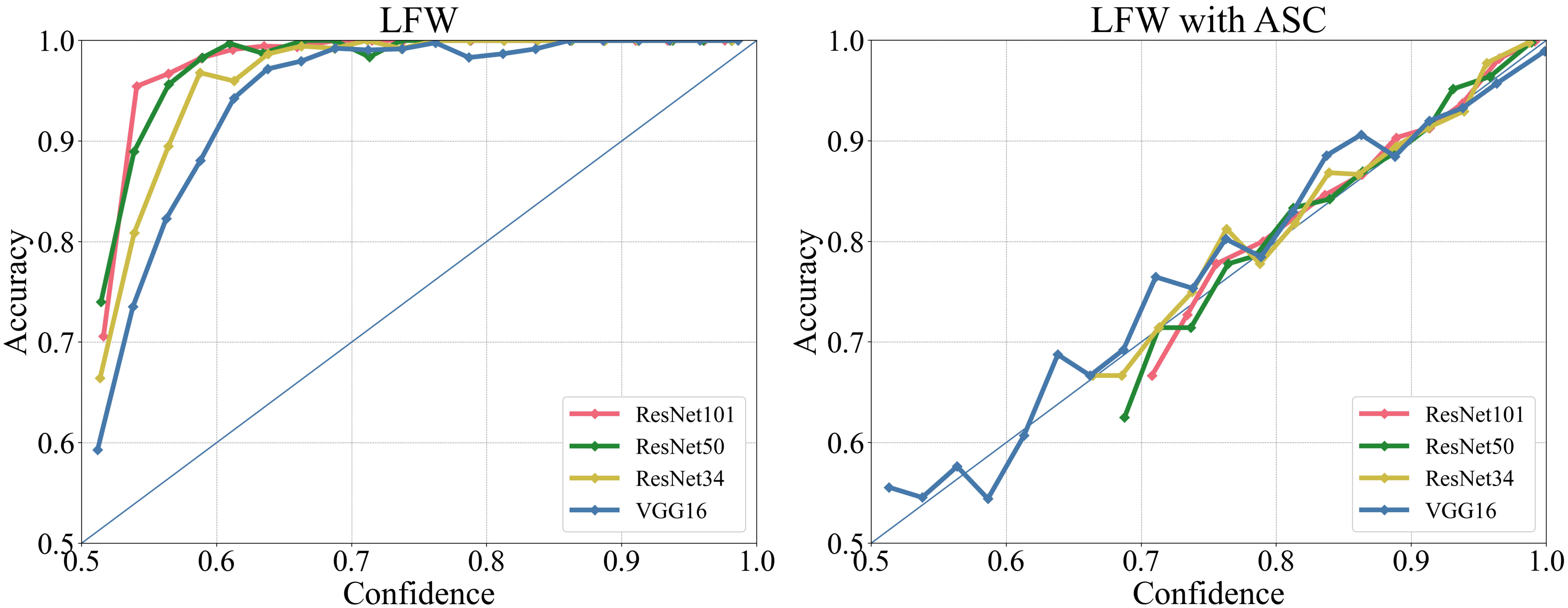}
        \caption{Calibration curves of different models before and after calibration (with ASC) on LFW \cite{huang2008labeled}.}
        \label{Fig6}
\end{minipage}\quad
\begin{minipage}[t]{0.48\textwidth}
        \vspace{0pt}
		\centering
		\makeatletter\def\@captype{figure}\makeatother
        \includegraphics[width=\linewidth]{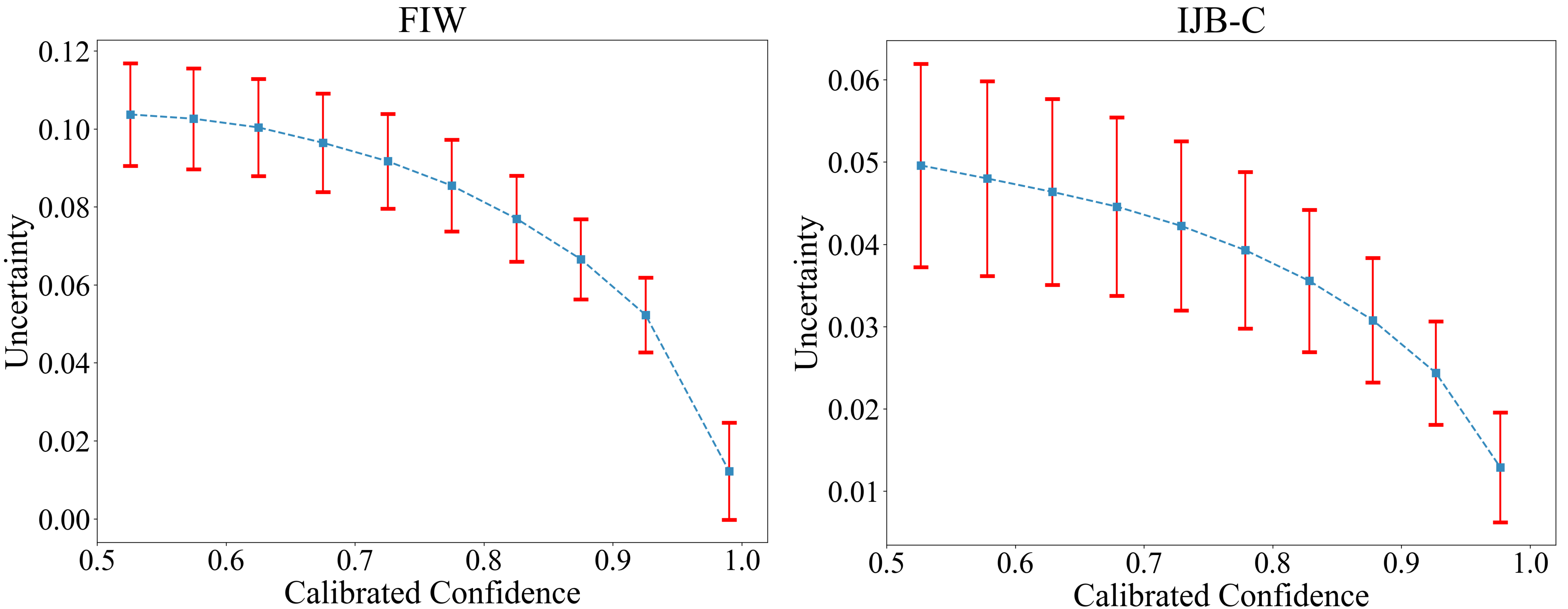}
        \caption{The trend of uncertainty vs. calibrated confidence on FIW \cite{robinson2016families} and IJB-C \cite{maze2018iarpa}.}
        \label{Fig7}
\end{minipage}
\end{figure*}

\subsection{Experimental Results on LFW}

To further evaluate our methods on face verification task, we also conduct experiments on the LFW dataset \cite{huang2008labeled}. As can be seen from Table \ref{tab:table5}, deeper architectures may achieve better verification accuracy, but the calibration performance gets worse, which has also been demonstrated on both FIW \cite{robinson2016families} and LFW \cite{huang2008labeled}. Moreover, Fig. \ref{Fig6} plots the correlation between the confidence and accuracy of the four models before and after calibration. We observe that ASC is helpful to improve the calibration of all four models. Due to the high accuracy of ResNet34, ResNet50, and ResNet101 on the LFW dataset \cite{huang2008labeled}, the ASC is inclined to map the prediction confidence to a high-level, consequently aligning with the expected accuracy. However, as shown in Fig. \ref{Fig6}, the calibrated confidence of all four models does not degenerate to a constant solution. Instead, it is effectively adjusted to match the prediction accuracy at varied levels.

\subsection{Experimental Results on IJB-C}

We present in Table \ref{tab:table6} the threshold $\tau$, the TAR (True Accept Rate), the accuracy, the mean confidence, and the ECE before calibration on IJB-C dataset \cite{maze2018iarpa}. The threshold increases as the FAR (False Accept Rate) drops from 1.0\% to 0.01\%, resulting in lower TAR and ECE. Upon evaluating results from four distinct groups of experiments, a consistent reduction in ECE is observed as the TAR decreases. This finding is consistent with the correlation between ECE and accuracy. Moreover, we further discover that deeper models, in the process of achieving a superior TAR, also exhibit a larger ECE. This finding aligns with observations drawn from preceding experiments, highlighting a interaction between model depth, TAR, and ECE. In addition, a focused analysis evaluating the impact of loss functions on ECE revealed significant distinctions. Specifically, the ResNet50 model, when trained using the Triplet loss\cite{schroff2015facenet}, demonstrated a higher ECE in comparison to the model trained with the ArcFace loss \cite{deng2019arcface}. Interestingly, it even exceeded the ECE of the more deeper ResNet101 model. This observation underscores the critical role that the selection of a loss function plays in affecting a model's calibration performance.

\begin{figure*}[!t]
        \centering
        \begin{minipage}[t]{0.43\textwidth}
        \vspace{0pt}
		\centering
		\makeatletter\def\@captype{table}\makeatother
        \renewcommand\arraystretch{1.0}
        \caption{\scriptsize Accuracy (\%), confidence (\%), and ECE (\%) of different models before and after calibration (with ASC) on LFW \cite{huang2008labeled}}
        \resizebox{\linewidth}{!}{
    	 \begin{tabular}{c|c|c|c|c|c}
            \bottomrule
            Model &Acc. ($\uparrow$)&Conf.&\makecell{Conf.\\w/ ASC}&ECE ($\downarrow$)&\makecell{ECE ($\downarrow$)\\w/ ASC} \\
            \hline
            ResNet101&\textbf{99.55}&70.65&98.79&28.48&\textbf{0.78}\\
            ResNet50 &99.23&71.68&98.33&27.84&0.93\\
            ResNet34 &98.88&73.14&97.81&25.43&1.13\\
            VGG16    &93.57&70.89&93.31&\textbf{24.67}&1.54\\
            \toprule
            \end{tabular}
        }
        \label{tab:table5}
        \end{minipage}
        \begin{minipage}[t]{0.5\textwidth}
        \vspace{0pt}
        \centering
        \makeatletter\def\@captype{table}\makeatother
        \setcounter{table}{6}
        \renewcommand\arraystretch{1.108}
        \caption{\scriptsize Accuracy (\%), confidence (\%), and ECE (\%) of different models after calibration (with ASC) on IJB-C \cite{maze2018iarpa}}
         \resizebox{\linewidth}{!}{
        \begin{tabular}{c|c|c|c|c|c|c|c}
        \bottomrule
        \multirow{2}{*}{Loss} & \multirow{2}{*}{Model} & \multicolumn {3}{c|}{FAR=1.0\%} & \multicolumn {3}{c}{FAR=0.01\%}  \\
        \cline{3-8}
        && Acc. ($\uparrow$)& Conf. & ECE ($\downarrow$)& Acc. ($\uparrow$)& Conf. &  ECE ($\downarrow$)\\
        \hline
        \multirow{3}{*}{ArcFace}& ResNet101 &\textbf{99.92}&98.77&1.30&\textbf{99.94}&99.52&\textbf{0.47} \\
        \cline{3-8}
                                & ResNet50  &\textbf{99.92}&98.80&\textbf{1.18}&99.93&99.46&0.53\\
        \cline{3-8}
                                & ResNet34  &99.89&98.06&1.90&99.93&99.36&0.63 \\
        \hline
        Triplet                 &ResNet50   &99.71&97.72&2.24&99.90&97.95&2.19 \\
        \toprule
        \end{tabular}
        }
        \label{tab:table7}
\end{minipage}\quad
\end{figure*}

\setcounter{table}{5}
\begin{table*}[!t]
\renewcommand\arraystretch{1.2}
\caption{ Accuracy (\%), confidence (\%), and ECE (\%) of different models before calibration on IJB-C \cite{maze2018iarpa}
\label{tab:table6}}
\scriptsize
\centering
\resizebox{0.94\linewidth}{!}{
    \begin{tabular}{c|c|c|c|c|c|c|c|c|c|c|c|c}
    \bottomrule
    \multirow{2}{*}{Loss} & \multirow{2}{*}{Model} & \multicolumn {5}{c|}{FAR=1.0\%} & \multicolumn {5}{c|}{FAR=0.01\%} & \multirow{2}{*}{AUC ($\uparrow$)} \\
    \cline{3-12}
    &&{\normalsize $\tau$} & TAR ($\uparrow$)& Acc. ($\uparrow$)& Conf. & ECE ($\downarrow$)& {\normalsize $\tau$} & TAR ($\uparrow$) & Acc. ($\uparrow$) & Conf. &  ECE ($\downarrow$)&  \\
    \hline
    \multirow{3}{*}{ArcFace}& ResNet101 & 0.03&\textbf{97.17}&\textbf{99.92}&65.54&34.89&0.22&\textbf{93.92}&\textbf{99.94}&70.76&29.64&\textbf{99.25} \\
    \cline{2-13}
                            & ResNet50 & -0.17&96.69&\textbf{99.92}&68.62&31.36&0.09&90.16&99.93&75.47&23.97&\textbf{99.25} \\
    \cline{2-13}
                            & ResNet34 & -0.20&96.02&99.89&71.99&\textbf{27.98}&0.02&90.56&99.93&77.56&\textbf{21.93}&99.24 \\
    \hline
    Triplet                  &ResNet50& 0.15&93.35&99.71&64.73&35.10&0.28&86.31&99.90&68.31&31.36&99.07 \\
    \toprule
    \end{tabular}
}
\end{table*}

\setcounter{table}{7}

\begin{table}[t]
\renewcommand\arraystretch{1.3}
  \centering
  \caption{Accuracy (\%), confidence (\%), and ECE (\%) of different losses and models before and after calibration (with ASC) in cross-dataset settings}
  \resizebox{\linewidth}{!}{
        \Large
        \begin{tabular}{c|c|c|c|c|c|c|c}
        \bottomrule
              & \multicolumn{1}{c|}{Loss} & Model & \multicolumn{1}{c|}{Acc ($\uparrow$)} & \multicolumn{1}{c|}{Conf.} & \multicolumn{1}{c|}{\makecell{Conf.\\w/ ASC}} & \multicolumn{1}{c|}{ECE ($\downarrow$)} & \multicolumn{1}{c}{\makecell{ECE ($\downarrow$) \\w/ ASC}} \\
        \hline
        \multicolumn{1}{c|}{\multirow{6}[3]{*}{\makecell{FIW \\ (KFW-II)}}} & \multicolumn{1}{c|}{\multirow{3}[1]{*}{ArcFace}} & ResNet50 & \textbf{74.95} & 58.59 & 70.81 & 16.36 & 5.18 \\
              &       & ResNet34 & 72.65 & 62.30  & 70.80  & 10.35 & 4.32 \\
              &       & VGG16 & 61.45 & 66.59 & 62.72 & \textbf{7.66}  & \textbf{3.05} \\
         \cline{2-8}          & \multicolumn{1}{c|}{\multirow{3}[2]{*}{InfoNCE}} & ResNet50 & \textbf{75.80}  & 59.57 & 74.54 & 16.23 & 4.48 \\
              &       & ResNet34 & 73.30  & 60.61 & 70.87 & 12.69 & 3.51 \\
              &       & VGG16 & 64.95 & 62.57 & 65.72 & \textbf{5.45}  & \textbf{2.71} \\
        \hline
        \multicolumn{1}{c|}{\multirow{4}[2]{*}{\makecell{IJB-C \\ (LFW)}}} & \multicolumn{1}{c|}{\multirow{4}[2]{*}{ArcFace}} & ResNet101 & \textbf{98.87} & 70.16 & 99.08 & 28.71 & 1.17 \\
              &       & ResNet50 & 98.82 & 68.36 & 99.39 & 28.45 & 1.05 \\
              &       & ResNet34 & 98.78 & 69.44 & 99.32 & 28.33 & \textbf{0.96} \\
              &       & VGG16 & 85.82 & 65.92 & 85.87 & \textbf{24.16} & 1.85 \\
        \toprule
        \end{tabular}%
    }
  \label{tab:table8}%
\end{table}%

Table \ref{tab:table7} presents the accuracy, the mean confidence, and the ECE with ASC on the IJB-C dataset \cite{maze2018iarpa}. Upon calibration, it is evident that the confidence closely mirrors the true accuracy and the ECE tends towards zero. These findings validate the effectiveness of our proposed ASC method, exhibiting good calibration performance across a range of model backbones and diverse loss functions. Significantly, the ASC maintains a minimized calibration error, demonstrating its stability even when the threshold values undergo adjustments.

\subsection{Experimental Results on Cross-Dataset Verification}

To further assess the generalization capability of ASC, we perform face and kinship verification under cross-dataset settings:

\begin{enumerate}
\item FIW(KFW-II): We calibrate the model on the FIW dataset \cite{robinson2016families}, and then conduct tests on the KinFaceW-II dataset \cite{lu2013neighborhood}.
\item IJB-C(LFW): We use the IJB-C dataset \cite{maze2018iarpa} for calibration, and then evaluate its performance on the LFW dataset \cite{huang2008labeled}.
\end{enumerate}

The experimental results of these cross-dataset verification are shown in Table \ref{tab:table8}. In this setting, the accuracy of all models decreases, and the prediction confidences before calibration are generally lower than the prediction accuracy, resulting in a significant ECE. However, after calibration, the prediction confidence more closely aligns with the prediction accuracy, thereby achieving ECE at a lower level. These results underscore the generalization ability and potential of ASC in cross-dataset verification settings, highlighting the feasibility of employing ASC in real-world task scenarios.

\begin{figure}[t]
  \centering
  \includegraphics[width=\linewidth]{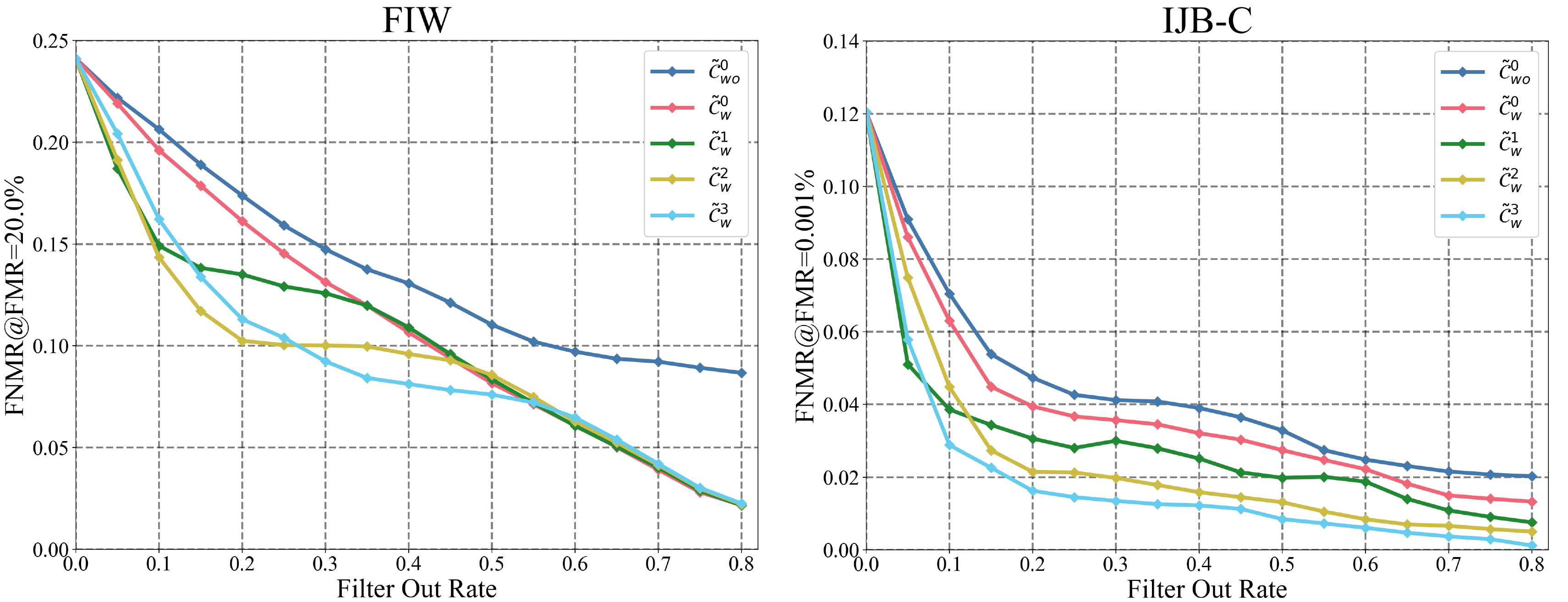}
  \caption{Error vs. Reject curves showcasing the incorporation of uncertainty estimation in prediction confidence on FIW \cite{robinson2016families} and IJB-C \cite{maze2018iarpa}.}
  \label{Fig8}
\end{figure}

\begin{figure}[t]
  \centering
  \includegraphics[width=\linewidth]{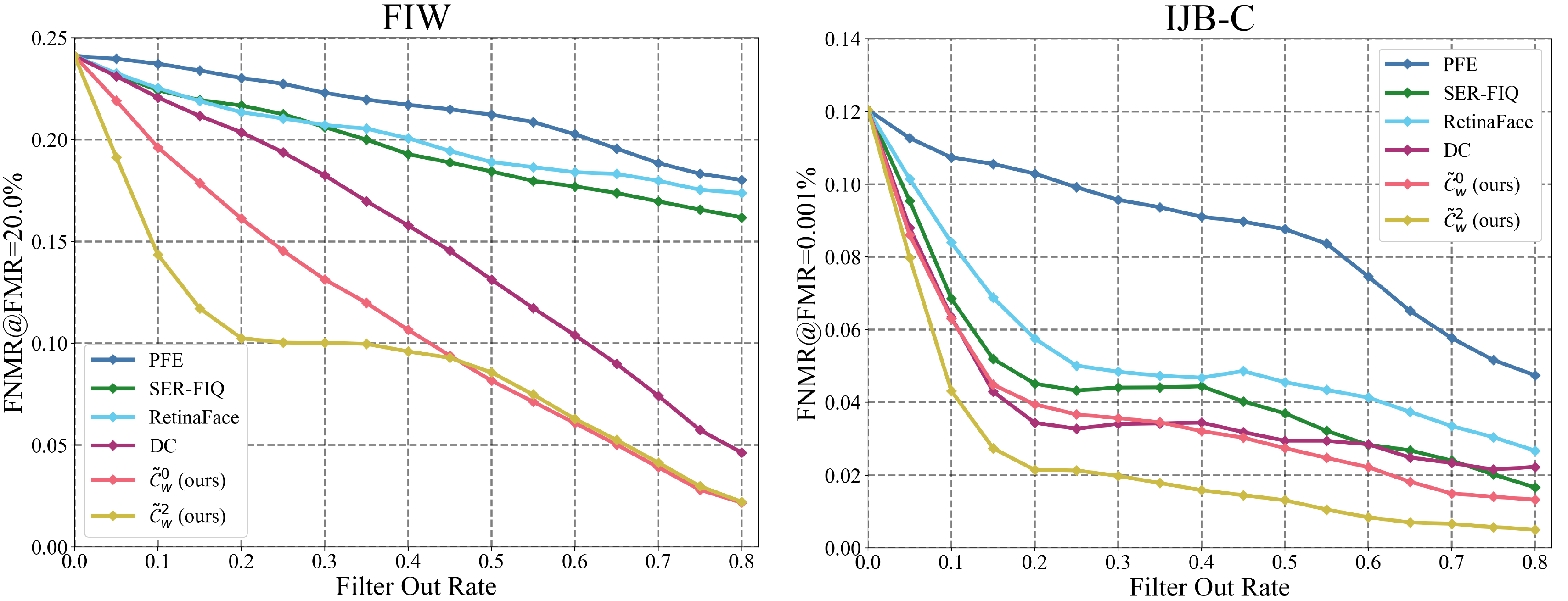}
  \caption{Error vs. Reject curves of our methods and alternative methods on FIW \cite{robinson2016families} and IJB-C \cite{maze2018iarpa}.}
  \label{Fig9}
\end{figure}

\setcounter{table}{9}

\begin{table*}[!t]
\renewcommand\arraystretch{1.126}
\caption{ Comparison of ECE (\%) for different post-calibration methods on four datasets. In each group, the best result is marked in red}
\label{tab:table10}
\tiny
\centering
  \resizebox{\linewidth}{!}{
\begin{tabular}{c|c|c|c|c|c|c}
\bottomrule
Dataset & Loss (Method) & Model (Feature) & Uncalibrated & Histogram binning\cite{zadrozny2001obtaining} & Isotonic regression\cite{zadrozny2002transforming}& ASC \\
\hline
\multirow{16}{*}{FIW} & \multirow{4}{*}{InfoNCE}&ResNet101&22.85&2.47&2.51&\textcolor{red}{2.08}\\
\cline{3-7}
&&ResNet50&16.51&\textcolor{red}{1.57}&1.62&1.81\\
\cline{3-7}
&&ResNet34&12.72&2.89&2.60&\textcolor{red}{2.11}\\
\cline{3-7}
&&VGG16&5.97&2.14&2.17&\textcolor{red}{1.39}\\
\cline{2-7}

& \multirow{4}{*}{Triplet}&ResNet101&20.93&\textcolor{red}{1.22}&1.36&2.83\\
\cline{3-7}
&&ResNet50&16.97&2.83&\textcolor{red}{2.82}&2.83\\
\cline{3-7}
&&ResNet34&10.46&2.80&2.57&\textcolor{red}{1.87}\\
\cline{3-7}
&&VGG16&7.87&\textcolor{red}{1.51}&4.70&1.54\\
\cline{2-7}
&\multirow{4}{*}{ArcFace}&ResNet101&19.39&3.14&2.95&\textcolor{red}{2.62}\\
\cline{3-7}
&&ResNet50&16.00&4.75&4.78&\textcolor{red}{2.45}\\
\cline{3-7}
&&ResNet34&10.04&3.45&3.76&\textcolor{red}{3.40}\\
\cline{3-7}
&&VGG16&7.69&1.66&\textcolor{red}{1.40}&2.90\\
\cline{2-7}
& \multirow{4}{*}{Softmax}&ResNet101&16.24&1.37&\textcolor{red}{1.20}&2.79\\
\cline{3-7}
&&ResNet50&15.77&2.92&2.85&\textcolor{red}{2.80}\\
\cline{3-7}
&&ResNet34&8.90&3.82&3.65&\textcolor{red}{3.56}\\
\cline{3-7}
&&VGG16&5.40&2.05&1.84&\textcolor{red}{1.81}\\
\hline

\multirow{3}{*}{KinFaceW-I}& \multirow{2}{*}{NRML} &LBP &20.95&\textcolor{red}{1.90}&3.06&2.14 \\
\cline{3-7}
&&HOG &22.30&4.98&\textcolor{red}{4.56}&4.59\\
\cline{2-7}
&InfoNCE& ResNet18& 27.13&1.69&2.92&\textcolor{red}{1.64}\\
\hline
\multirow{3}{*}{KinFaceW-II}& \multirow{2}{*}{NRML} &LBP &19.38&\textcolor{red}{1.08}&1.34&1.53\\
\cline{3-7}
&&HOG &20.78&2.22&\textcolor{red}{1.37}&1.62\\
\cline{2-7}
&InfoNCE& ResNet18& 26.56&2.35&2.22&\textcolor{red}{1.00}\\

\hline
\multirow{4}{*}{LFW}& \multirow{4}{*}{ArcFace}&ResNet101&28.48&1.22&\textcolor{red}{0.16}&0.78\\
\cline{3-7}
&&ResNet50&27.84&0.98&\textcolor{red}{0.70}&0.93\\
\cline{3-7}
&&ResNet34&25.43&2.35&1.76&\textcolor{red}{1.13}\\
\cline{3-7}
&&VGG16&24.67&2.56&1.58&\textcolor{red}{1.54}\\
\hline
\multirow{4}{*}{IJB-C}& \multirow{3}{*}{ArcFace}&ResNet101&29.64&1.18&0.97&\textcolor{red}{0.47} \\
\cline{3-7}
&&ResNet50&23.97&1.49&1.00&\textcolor{red}{0.53}\\
\cline{3-7}
&&ResNet34&21.93&1.72&1.14&\textcolor{red}{0.63}\\
\cline{2-7}
&Triplet&ResNet50&31.36&2.24&\textcolor{red}{2.07}&2.19\\
\toprule
\end{tabular}
}
\end{table*}

\setcounter{table}{8}
\begin{table}[t]
  \centering
  \renewcommand\arraystretch{1.2}
  \caption{ECE (\%) of prediction confidence incorporating uncertainty estimation on FIW \cite{robinson2016families} and IJB-C \cite{maze2018iarpa}}
  \resizebox{0.7\linewidth}{!}{
    \begin{tabular}{c|ccccc}
    \bottomrule
    \multicolumn{1}{c|}{} &$\tilde{\mathcal{C}}_{wo}^{0}$  & $\tilde{\mathcal{C}}_{w}^{0}$  & $\tilde{\mathcal{C}}_{w}^{1}$  & $\tilde{\mathcal{C}}_{w}^{2}$ & $\tilde{\mathcal{C}}_{w}^{3}$  \\
    \hline
    FIW   & 19.31 & 2.59  & \textbf{2.10}  & 2.27  & 3.27 \\
    IJB-C & 27.51 & \textbf{0.45}  & 0.98  & 1.36  & 1.72 \\
    \toprule
    \end{tabular}%
  }
  \label{tab:table9}%
\end{table}%

\setcounter{table}{10}

\subsection{Experimental Results on Uncertainty Estimation}

To validate the effectiveness of our proposed uncertainty estimation method, we conduct experiments on the FIW \cite{robinson2016families} and the IJB-C \cite{maze2018iarpa} datasets. An additional network branch is incorporated into a ResNet101 model trained using the ArcFace loss \cite{deng2019arcface}, with the objective of learning to predict confidence uncertainty $\sigma_c$ as determined by Eq. \eqref{equation_11}. To facilitate comparison, the backbone was frozen during the training process.

Fig. \ref{Fig7} illustrates the relationship between uncertainty $\sigma_c$ and calibrated confidence. From a statistical viewpoint, we note a decrease in uncertainty as prediction confidence heightens, suggesting an overall enhancement in image quality. To compare and analyze the impact of uncertainty estimation on decision-making, we set up the following prediction confidence schemes: $\tilde{\mathcal{C}}_{wo}^{0}$, $\tilde{\mathcal{C}}_{w}^{0}$, $\tilde{\mathcal{C}}_{w}^{1}$, $\tilde{\mathcal{C}}_{w}^{2}$, and $\tilde{\mathcal{C}}_{w}^{3}$. In these notations, the subscript $w$ signifies the utilization of ASC, while $wo$ indicates without the use of ASC; the superscript represents the value of the uncertainty weighting factor $\alpha$ in Eq. \eqref{equation_14}. Fig. \ref{Fig8} displays the Error vs. Reject curves \cite{grother2007performance} for the five aforementioned schemes, in which the error rate progressively diminishes as more samples of low-confidence are excluded, indicating the effectiveness of our proposed confidence measure method. Moreover, compared to $\tilde{\mathcal{C}}_{wo}^{0}$, the use of calibrated confidence facilitates a more rapid reduction in error rate. The introduction of uncertainty estimation additionally contributes to a lower decision error rate, thereby demonstrating the efficacy of our proposed methods for confidence calibration and uncertainty estimation.

We compare the decision performance of our proposed methods ($\tilde{\mathcal{C}}_{w}^{0}$ and $\tilde{\mathcal{C}}_{w}^{2}$) with alternative methods, as shown in Fig. \ref{Fig9}. These methods include:

\begin{figure*}[!t]
\centering
\includegraphics[width=0.965\textwidth]{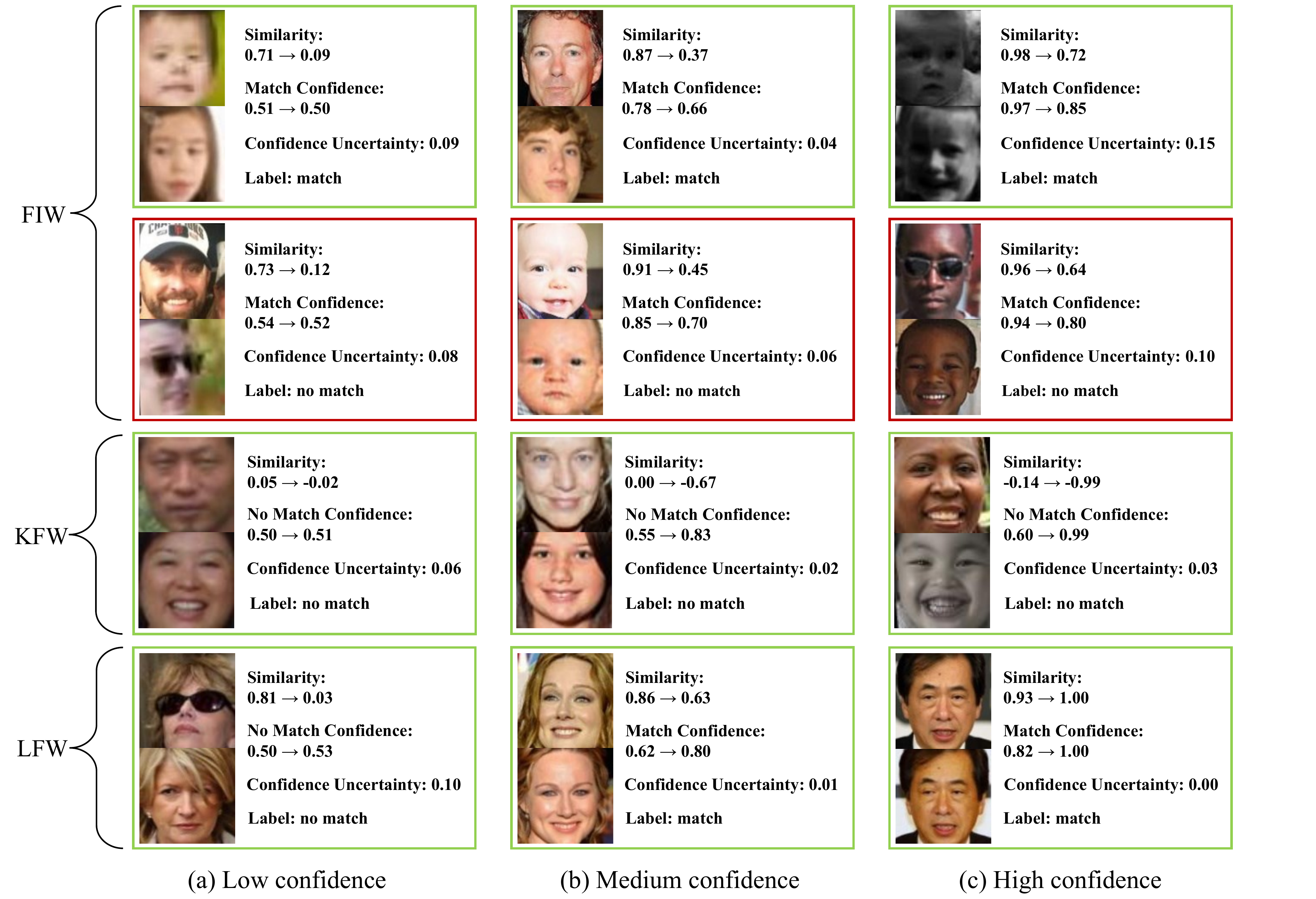}
\centering
\caption{
    Visualization of prediction confidence by our proposed method. From top to bottom are examples from FIW \cite{robinson2016families}, KFW \cite{lu2013neighborhood}, and LFW \cite{huang2008labeled} datasets, respectively. Within each group, we demonstrate the similarity and confidence before and after calibration, alongside the confidence uncertainty. To facilitate a comprehensive visualization, we exhibit sample pairs categorized into three distinct confidence levels: \textbf{low} confidence (a), \textbf{medium} confidence (b), and \textbf{high} confidence (c). Instances of misclassification are highlighted by red boxes.}
\label{Fig10}
\end{figure*}

\begin{enumerate}
\item PFE \cite{shi2019probabilistic}: This method employs data uncertainty for the estimation of face image quality, using the inverse of the harmonic mean of the uncertainty in each face embedding dimension as the image quality metric.
\item SER-FIQ \cite{terhorst2020ser}: This method utilizes the Bayesian approximation of model uncertainty to evaluate face image quality, and we use the setting of SER-FIQ (same model) \cite{terhorst2020ser} to compare with other methods.
\item RetinaFace \cite{deng2020retinaface}: The confidence score derived from RetinaFace face detection is used as the face image quality metric.
\item DC \cite{huber2022stating}: This method uses model uncertainty and the uncertainty propagation formula \cite{ku1966notes} to estimate decision confidence, and we set $\alpha=2$ following \cite{huber2022stating}.
\end{enumerate}

PFE \cite{shi2019probabilistic}, SER-FIQ \cite{terhorst2020ser}, and RetinaFace \cite{deng2020retinaface} achieve suboptimal decision performance, as they solely focus on image quality and neglect the verification process. In contrast, our proposed confidence measure and confidence calibration methods demonstrate superior decision performance compared to DC \cite{huber2022stating}. Additionally, our method also offers a computational efficiency without the need for expensive Monte Carlo Dropout \cite{gal2016dropout}.

Furthermore, we examine the calibration performance of the five schemes, as shown in Table \ref{tab:table9}. Compared to $\tilde{\mathcal{C}}_{w}^{0}$, the integration of uncertainty estimation brings about some degradation in calibration performance. However, the calibration error remains at a low level. For the FIW \cite{robinson2016families} and the IJB-C \cite{maze2018iarpa} datasets, setting $\alpha$ equal to 2 is a optimal choice to balance the decision-making and the calibration performance.

\subsection{Comparison with Previous Post-calibration Methods}

To validate the superiority of our ASC for confidence calibration in face and kinship verification, two widely adopted post-calibration methods are chosen for performance comparison: Histogram binning \cite{zadrozny2001obtaining} and Isotonic regression \cite{zadrozny2002transforming}. While there are some other post-calibration methods, such as Temperature scaling \cite{guo2017calibration}, Platt scaling \cite{platt1999probabilistic}, Bayesian binning \cite{naeini2015obtaining}, and Matrix and vector scaling, they are not directly applicable to confidence calibration for face and kinship verification tasks. For fair comparison, we maintain the same experimental setting across all compared methods.

\textbf{Histogram binning} \cite{zadrozny2001obtaining} is a non-parametric calibration method. In this paper, Histogram binning firstly divide the cosine similarity score $s$ evenly into $M$ bins $B_1$, $B_2$, ..., $B_M$, with bin boundaries $-1=a_{1}<a_{2}<...<a_{M}<a_{M+1}=1$, where the bin $B_m$ refers to $\left( a_{m}, a_{m+1}\right]$. We assign each $B_m$ a calibration score $\eta_m$, which is optimized by:
\begin{equation}
\label{equation_15}
\mathop{\arg\min}\limits_{\eta_{1},...,\eta_{M}}\sum\limits_{m=1}^{M}\sum\limits_{i=1}^{N} 1(a_{m}<s_{i} \leq a_{m+1})(\eta_{m}-Z_{i})^2
\end{equation}
where $s_i$, $Z_{i}$ are uncalibrated similarity score and the label of the $i$th sample pair, respectively. Once the $i$th sample falls into bin $B_m$, its calibrated similarity will be set to $\eta_m$. Additionally, the calibrated threshold $\tau'$ is updated with $\eta_t$, where $\tau \in \left( a_{t}, a_{t+1}\right]$.

\textbf{Isotonic regression} \cite{zadrozny2002transforming} is another commonly used non-parametric calibration method that can calibrate similarity score by learning a piecewise constant function $\mathcal{I}$, that is, $s'=\mathcal{I}(s)$, where $\mathcal{I}(s_{i})\leq \mathcal{I}(s_{j})$ if $s_{i} \leq s_{j}$. The goal of Isotonic regression is to minimize the square loss of $\sum_{i=1}^{N}(\mathcal{I}(s_{i})-Z_{i})^2$, which is optimized to learn the boundaries and calibration scores of each interval. We employ PAVA \cite{de2010isotone} algorithm to optimize the function $\mathcal{I}$, and the calibrated threshold $\tau'$ is set to $\mathcal{I}(\tau)$.

Table \ref{tab:table10} shows the calibration of three alternative methods (ASC, Histogram binning \cite{zadrozny2001obtaining}, and Isotonic regression \cite{zadrozny2002transforming}) on the FIW \cite{robinson2016families}, KinFaceW \cite{lu2013neighborhood}, LFW \cite{huang2008labeled}, and IJB-C \cite{maze2018iarpa} datasets, utilizing the ECE metric. We observe that ASC achieves the best calibration performance in most cases. ASC is easy to implement, as it involves only two parameters to be optimized and they can be efficiently learned by most gradient-based optimizers. In addition, ASC has a differentiable calibration function, making it seamlessly integrable with uncertainty estimation.

\subsection{Visualization Analysis}

Fig. \ref{Fig10} illustrates prediction confidence on sampled pairs from three datasets. These pairs are categorized into three groups based on the calibrated confidence: low confidence, medium confidence, and high confidence. The visualization results indicate that ASC can appropriately increase or decrease prediction confidence to match the true accuracy, providing accurate confidence estimation for system decision-making. In addition, samples with low confidence are usually accompanied by certain facial occlusions or image blurring. However, there are also some samples that lack facial information but have high confidence (e.g., the case in the top right). The application of uncertainty estimation enables us to adjust prediction confidence by taking into account both image quality and the image similarity, thereby bolstering the reliability and trustworthiness of the verification system.

\section{Conclusions}
\label{Section5}
In this work, we propose a simple yet effective confidence measure for face and kinship verification tasks, allowing any off-the-shelf verification models to estimate the decision confidence in an efficient and flexible way. We further introduce a confidence calibration method using angular scaling for face and kinship verification, which is retraining-free and accuracy-preserving. Incorporating our proposed confidence measure and confidence calibration methods with uncertainty estimation, we succeed in jointly model the image quality and verification process. This integration results in a notable improvement in decision-making performance while maintaining calibration performance. Comprehensive experiments are conducted on four widely used face and kinship verification datasets to examine the calibration of prevalent facial and kinship verification models. The effectiveness of our proposed methods is validated through these experiments. A comparative analysis with two popular post-calibration methods shows that our ASC exhibits superior calibration performance. In future research, we will explore improving the generalization ability and versatility of our approach, adapting it to diverse application domains such as medical imaging or the manufacturing industry.

\bibliographystyle{IEEEtran}
\bibliography{main}

\clearpage

\renewcommand*{\theequation}{S\arabic{equation}}
\setcounter{equation}{0}

\renewcommand{\appendixname}{Appendix}

\appendix
\label{appendix}
We present a detailed derivation procedure for attaining uncertainty estimation in confidence calibration. For a pair of face images $(X, Y)$ and their corresponding face embeddings $(x, y)$, as well as the uncertainty (variance) in the features $(\sigma_x^2, \sigma_y^2)$, where $\vert \vert x \vert \vert, \vert \vert y \vert \vert=1$, the cosine similarity can be written as:
\begin{equation}
\label{equationS1}
s(x,y)=\frac{x \cdot y}{\vert\vert x\vert\vert \thinspace \vert\vert y \vert\vert}= x \cdot y
\end{equation}
According to the formula of uncertainty propagation \cite{ku1966notes},
\begin{equation}
\label{equationS2}
\sigma_{\mathcal{F}(\mathbf{x})}^2=\Big (\frac{\partial \mathcal{F}}{\partial \mathbf{x}}\Big )^2\sigma_{\mathbf{x}}^2
\end{equation}
where $\mathcal{F}$ is an arbitrary differentiable function with respect to $\mathbf{x}$. The uncertainty of the similarity $s$ is:
\begin{gather}
\label{equationS3}
\begin{align}
\sigma_{s}^2&=\Big( \frac{\partial s}{ \partial x} \Big)^2\sigma_{x}^2+\Big( \frac{\partial s}{ \partial y} \Big)^2\sigma_{y}^2 \nonumber \\
            &=\sum_{l=1}^Ly_l^2\sigma_{x_l}^2+\sum_{l=1}^Lx_l^2\sigma_{y_l}^2
\end{align}
\end{gather}
where $L$ is the number of feature dimension. Then we define the calibration function as:
\begin{equation}
\label{equationS4}
\psi(s)=cos(arccos(s)*w+b)
\end{equation}
The calibrated similarity and threshold are denoted as: $s'=\psi(s), \tau'=\psi(\tau)$, and the uncertainty of the calibrated similarity is:
\begin{gather}
\label{equationS5}
\begin{align}
\sigma_{s'}^2&=\Big( \frac{\partial s'}{ \partial s} \Big)^2\sigma_s^2 \nonumber \\
             &= \frac{w^2(1-\psi^2(s))}{1-s^2}   \sigma_s^2
\end{align}
\end{gather}
Specifically, when $w=1$ and $b=0$, we have $\sigma_{s'}^2=\sigma_{s}^2$. For confidence measure, we denote it as:
\begin{equation}
\label{equationS6}
c(s,\tau)=\left\{
\begin{aligned}
    \frac{1}{2} \thinspace \frac{s-\tau}{1-\tau}+\frac{1}{2}, & \qquad s \geq \tau \\
    \frac{1}{2} \thinspace \frac{\tau-s}{1+\tau}+\frac{1}{2}, & \qquad s < \tau \\
\end{aligned}
\right.
\end{equation}
The uncertainty of the calibrated confidence $c$ is:
\begin{equation}
\label{equationS7}
\sigma_{c}^2=\Big( \frac{\partial c}{ \partial s'} \Big)^2\sigma_{s'}^2=\left\{
             \begin{aligned}
                \frac{1}{4} \thinspace \frac{1}{(1-\tau')^2}\sigma_{s'}^2, & \qquad s' \geq \tau' \\
                \frac{1}{4} \thinspace \frac{1}{(1+\tau')^2}\sigma_{s'}^2, & \qquad s' < \tau' \\
             \end{aligned}
                \right.
\end{equation}

Combining Eq. \eqref{equationS3}, \eqref{equationS5}, and \eqref{equationS7}, the uncertainty of calibrated confidence can be expressed as:

\begin{equation}
\label{equationS8}
\sigma_c^2(x,y,\sigma_x^2,\sigma_y^2)=\left\{
\begin{aligned}
    \frac{\sigma_s^2(x,y,\sigma_x^2,\sigma_y^2)}{4[1-\psi(\tau)]^2}, & \quad s(x,y)\geq \tau \\
    \frac{\sigma_s^2(x,y,\sigma_x^2,\sigma_y^2)}{4[1+\psi(\tau)]^2}, & \quad s(x,y) < \tau \\
\end{aligned}
\right.
\end{equation}
where
\begin{equation}
\label{equationS9}
\sigma_s^2(x,y,\sigma_x^2,\sigma_y^2)=\frac{w^2[1-\psi^2[s(x,y)]]}{1-s^2(x,y)}\sum_{l=1}^{L}(y_l^2\sigma_{x_l}^2+x_l^2\sigma_{y_l}^2)
\end{equation}
When $w=1$ and $b=0$ in Eq. \eqref{equationS8}, the resulting calculation represents the uncertainty of uncalibrated confidence.

\end{document}